\documentclass{article}

\usepackage[main,final]{neurips_2026}

\usepackage[utf8]{inputenc} 
\usepackage[T1]{fontenc}    
\usepackage[hidelinks]{hyperref}       
\usepackage{url}            
\usepackage{booktabs}       
\usepackage{amsfonts}       
\usepackage{nicefrac}       
\usepackage{microtype}      
\usepackage{xcolor}         
\usepackage{graphicx}
\usepackage{caption}
\usepackage{subcaption}
\usepackage[normalem]{ulem}
\usepackage{amsmath}
\usepackage{amssymb}
\usepackage{mathtools}
\usepackage{array, makecell}
\usepackage{multirow}
\usepackage{booktabs}
\usepackage{wrapfig}
\usepackage{tabularx}
\usepackage{pifont}
\newcommand{\cmark}{\ding{51}}
\newcommand{\xmark}{\ding{55}}

\title{SensOFormer: Foundation Models for Particle Accelerators}

\author{
  Mahindra Rautela, Alexander Scheinker \\[0.5em]
  Los Alamos National Laboratory, Los Alamos, NM, USA. \\
  \texttt{\{mrautela,ascheink\}@lanl.gov}\\
}

\begin{document}

\maketitle

\begin{abstract}
  Large-scale scientific facilities, such as particle accelerators, are complex systems composed of thousands of interacting components that must operate collectively to support diagnostics and control, and ensure safe, reliable operation. These facilities rely on extensive sensor networks that generate large volumes of heterogeneous, incomplete, and noisy telemetry, often sampled at different rates across subsystems. In this paper, we adopt the foundation-model paradigm to learn general-purpose sensor representations from historical accelerator data through self-supervised pretraining. We introduce SensOFormer, a denoising masked transformer with a Perceiver-style encoder–decoder architecture designed to handle variable sensor sets, accommodate heterogeneous sampling rates, and learn robust representations from noisy measurements. The model is pretrained on multiple particle-accelerator datasets and transferred to downstream tasks, including missing-data imputation, anomaly detection, cavity identification, and fault identification. Through extensive evaluations and comparisons, we demonstrate that a single self-supervised model can learn reusable representations for diverse operational tasks in large-scale scientific facilities. \textbf{Project page}: \url{https://mahindrautela.github.io/sensoformer}.
\end{abstract}

\section{Introduction} \label{sec:intro}

\textbf{Particle accelerators} are large-scale scientific facilities that enable fundamental studies in particle and nuclear physics, materials characterization using neutron and X-ray scattering, ultrafast imaging of chemical and biological processes, isotope production, radiation-effects testing, and high-energy-density science. Delivering these capabilities requires thousands of tightly coupled components to operate under strict stability, safety, and reliability constraints. Modern accelerators are therefore instrumented with extensive sensors and control systems that continuously monitor beam conditions, radio-frequency systems, magnets, vacuum, cooling, timing, and other subsystems \citep{chao2023handbook}. \emph{These instruments generate massive streams of operational telemetry that are noisy, incomplete, heterogeneous, and distributed across the facility}. As a result, core operational tasks such as missing-data recovery, anomaly detection, machine tuning, and control decision support depend critically on learning useful structure from high-dimensional sensor measurements \citep{edelen2024machine}.

\textbf{A foundation model to learn generalized sensor representations.}
The abundance of accelerator telemetry motivates a shift from training separate task-specific models for individual diagnostics, tuning, or control problems to learning general-purpose representations from large-scale unlabeled operational data. Foundation models are pretrained on broad, largely unlabeled data using self-supervised objectives and then adapted to many downstream tasks \citep{bommasani2021opportunities}. This recipe has driven progress in language modeling \citep{devlin2019bert,radford2019language}, masked visual representation learning \citep{he2022masked}, time-series modeling \citep{das2024decoder,ansari2024chronos}, and physical spatiotemporal modeling \citep{bodnar2025foundation,rautela2025morph}. Particle accelerators present an analogous opportunity: years of archived process variables, control signals, and beam diagnostics encode recurring machine states, subsystem couplings, operational drifts, and rare failure modes, yet most of these data are weakly labeled or unlabeled \citep{edelen2024machine}. \emph{A self-supervised accelerator foundation model can therefore turn routine operational data into a learning signal, producing reusable representations for pressing accelerator problems such as imputation, anomaly detection, cavity identification, and fault identification}. Task-specific models remain suitable for fixed sensor configurations with sufficient labeled data. However, a shared pretrained backbone enables lightweight adaptation across diagnostic tasks, reducing repeated model development and potentially improving label efficiency when annotated fault data are scarce.

\textbf{Challenges with operational data.}
Accelerator telemetry differs substantially from curated multivariate time-series benchmarks \citep{li2023timae}. Operational data are inherently \emph{heterogeneous}, collected from many subsystem-specific diagnostics, including superconducting RF cavity waveforms, RF station phase and amplitude signals, beam-position monitors, beam-loss monitors, magnet and power-supply variables, and high-voltage converter modulator waveforms \citep{tennant2020superconducting,st2021real,krymova2023data,liang2025coincident,radaideh2022time}. Each data acquisition (DAQ) segment may contain hundreds to thousands of channels, and at facility scale the number of available process variables can be much larger \citep{st2021real}. Moreover, the available sensors, their \emph{ordering}, and their \emph{grouping} can vary across sectors, subsystems, and acquisition periods. Consequently, the \emph{semantic identity} of a channel is determined by sensor metadata, physical location, and subsystem membership rather than by its column index \citep{krymova2023data}. Models that assume a fixed channel layout can therefore be brittle under sensor reordering, missing channels, or newly added diagnostics. Padding each acquisition to a facility-wide maximum channel set is also computationally inefficient at this scale. Second, accelerator measurements are \emph{asynchronous}: sensors are sampled at different rates, logged over variable-length windows, and interrupted by dropouts, maintenance periods, and DAQ-specific failures \citep{tennant2020superconducting}. Third, accelerators are \emph{time-varying systems} whose operating distributions drift with beam mode, machine configuration, component aging, calibration changes, and seasonal or operational conditions \citep{scheinker2023adaptive}. Finally, the data are \emph{noisy} with noise statistics that are difficult to specify a priori \citep{liang2025coincident}. Taken together, these properties make fixed-dimensional, task-specific modeling pipelines difficult to scale and prone to poor generalization across sensors, subsystems, and operating conditions. Instead, accelerator telemetry calls for models that can ingest variable numbers of sensor observations without padding, treat channels as an unordered or metadata-indexed set, handle missing and irregularly sampled measurements, and learn denoising representations directly from large-scale operational data.

\textbf{Model design requirements and SensOFormer.}
In light of the above-mentioned challenges, we design both the foundation model architecture and the pretraining strategy to accommodate heterogeneous accelerator telemetry. \emph{SensOFormer is a latent transformer model with a set encoder and set decoder}. The set encoder, inspired by perceiver-style models \citep{jaegle2021perceiver}, is \emph{permutation-invariant} across channels, allowing the model to process variable sensor sets without relying on fixed column semantics \citep{lee2019set}. The set decoder is \emph{permutation-equivariant} across channels, preserving the batch-specific sensor locations and identities of the queried outputs. Together, the encoder, latent transformer, and decoder provide flexibility across both channels and sequence lengths without requiring data or activation padding. This design is important for scaling to large accelerator datasets, where padding to the maximum number of sensors and time steps would be computationally inefficient. Finally, transformer-based architectures provide a scalable representation-learning framework, and the masked pretraining objective encourages the model to learn robust structure from noisy observations rather than overfitting to sensor-specific noise \citep{park2022vision, wang2022anti, batson2019noise2self,li2023timae}.

\paragraph{Contributions.}
We make the following contributions with this research:
\begin{enumerate}
    \item We introduce SensOFormer, to the best of our knowledge the \emph{first foundation model designed specifically for particle-accelerator operational telemetry}. 

    \item We pretrain SensOFormer on a diverse multi-facility benchmark spanning LANSCE, LHC, SNS, LCLS, and CEBAF datasets. These datasets cover operational telemetry, beam-loss monitoring, RF station signals, high-voltage converter modulator waveforms, and superconducting RF cavity fault data.

    \item We characterize scaling behavior with respect to model capacity and pretraining data volume. The scaling study shows that validation reconstruction loss improves with both larger models and more data, and provides guidance for compute-efficient scaling of future accelerator foundation models.

    \item We demonstrate strong zero-shot missing-data imputation across heterogeneous accelerator datasets, including transfer to a different year dataset (LANSCE-2025) and unseen accelerator facility (GMPS-Booster-Fermilab). We further show that pretrained SensOFormer representations can be adapted with lightweight task-specific heads for anomaly detection, cavity identification, and fault identification.
\end{enumerate}

\section{Related Work} \label{sec:related_works}

\paragraph{Scientific ML in particle accelerators.}
Machine learning has become an increasingly important tool for particle accelerators, with applications in beam-dynamics modeling, diagnostics, optimization, control, tuning, physics-constrained simulation, and uncertainty quantification \citep{rautela2024conditional,edelen2024machine,scheinker2025latent,scheinker2025physics,saxena2025improved,roussel2021multiobjective,edelen2020machine,wan2025symplectic,leon2024physics,mishra2021uncertainty,garcia2024machine,rautela:napac2025-mop002}. In operational settings, ML has enabled adaptive virtual diagnostics at Los Alamos Neutron Science Center (LANSCE at Los Alamos National Laboratory) \citep{scheinker2021adaptive,scheinker2025latent}, superconducting RF cavity-fault localization and classification at Continuous Electron Beam Accelerator Facility (CEBAF at Jefferson Laboratory) \citep{tennant2020superconducting}, anomaly detection in high-voltage converter modulator signals at Spallation Neutron Source (SNS at Oak Ridge National Laboratory) \citep{radaideh2022time}, beam-loss prediction at the Large Hadron Collider (LHC at CERN) \citep{krymova2023data}, RF station fault identification at Linac Coherent Light Source (LCLS at Stanford Linear Accelerator Laboratory) using beam-position-monitor and high-frequency RF phase data \citep{humble2022beam,liang2025coincident}, and accelerator control at Fermilab \citep{st2021real}. While these efforts show the effectiveness of specialist ML models, accelerator facilities share recurring RF, magnet, vacuum, timing, diagnostic, and control subsystems monitored by related sensor classes. SensOFormer leverages common operational and subsystem organization across facilities by learning transferable accelerator sensor representations through self-supervised foundation-model pretraining, rather than training task-specific models.

\paragraph{Scientific foundation models.}
SensOFormer adopts the foundation-model paradigm to learn reusable sensor representations from accelerator operational telemetry. This paradigm has been established in language modeling \citep{devlin2019bert,radford2019language} and computer vision \citep{he2022masked,caron2021emerging}, and has recently gained momentum across scientific domains, including physical systems \citep{rautela2025morph,park2025fm4npp,nguyen2023climax}, chemical science \citep{wadell2025foundation}, biological science \citep{moor2023foundation,brixi2026genome}, and time-series prediction \citep{ansari2024chronos,nie2022time}. Across these domains, foundation models succeed by scaling self-supervised pretraining over large, diverse, and heterogeneous datasets, while using input representations flexible enough to accommodate missing, irregular, or modality-specific structure. Accelerator telemetry exhibits these same properties: it is abundant but weakly labeled, heterogeneous across subsystems and facilities, and corrupted by noise and affected by changing operating conditions. SensOFormer brings the foundation-model paradigm to this setting through denoising masked reconstruction over multivariate sensor sets, enabling transferable representations for downstream tasks such as missing-data imputation, anomaly detection, cavity and fault identification. A detailed comparison of SensOFormer with existing set-based and any-variate time-series models, including SeFT \citep{horn2020set} and MOIRAI \citep{woo2024unified}, is provided in Appendix~\ref{app:related_works}.

\section{Proposed Method} \label{sec:methods}

\begin{figure}
\centering
\includegraphics[width=1.0\textwidth,trim=0cm 0cm 0cm 0cm, clip]{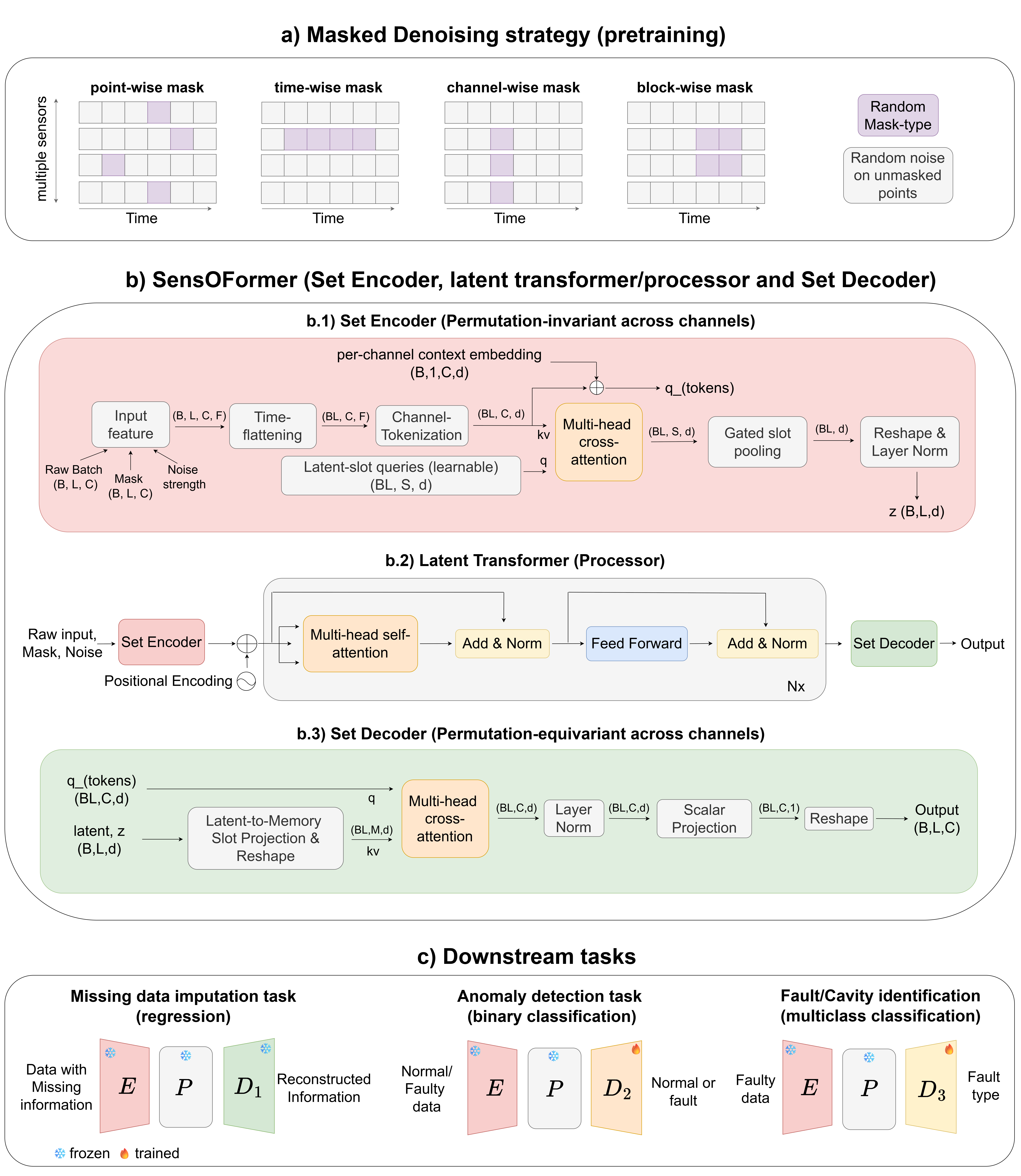}
\caption{\textbf{Overview of SensOFormer.} (a) Self-supervised pretraining uses masked denoising reconstruction. A mask is sampled from a mask bank containing point-wise, time-wise, channel-wise, and block-wise masks, while noise is added to the unmasked values. (b) SensOFormer uses an encoder--processor--decoder architecture for variable sensor sets. The perceiver-style set encoder is permutation-invariant across sensor channels. The latent transformer processor models temporal dependencies. The set decoder is permutation-equivariant, producing channel-wise reconstructions that preserve the queried sensor identities. (c) The pretrained encoder--processor backbone is reused with task-specific heads for missing-data imputation, anomaly detection, and cavity/fault identification.}
\label{fig:sensoformer}
\end{figure}

\subsection{Pretraining strategy}

Before describing the model architecture, we first motivate the pretraining objective. Foundation models are commonly pretrained using either autoregressive prediction, such as next-token, next-step, or next-frame prediction \citep{radford2019language,ansari2024chronos,rautela2025morph}, or masked reconstruction and denoising objectives \citep{devlin2019bert,nie2022time,wadell2025foundation}. Autoregressive forecasting is a natural pretraining choice when the primary goal is to extrapolate a sequence forward in time, as in language modeling, time-series forecasting for financial, energy, or traffic data, and physical systems with temporally evolving fields such as atmosphere and climate. Accelerator operational telemetry, however, does not generally form a single uniform sequence governed by one dominant time scale. Beam-related signals can correspond to fast machine events, since charged-particle bunches travel near the speed of light, whereas other subsystems, such as cooling, vacuum, and power-supply systems, evolve over slower operational time scales. Moreover, a central downstream task in this work is to recover missing or corrupted sensor measurements, rather than only to extrapolate future time steps. These considerations make masked denoising reconstruction a more natural pretraining objective for accelerator telemetry: the model must infer hidden measurements from the available temporal and cross-channel context while learning representations robust to noise, missingness, and heterogeneous sensor structure.

We therefore \emph{pretrain SensOFormer using a denoising masked reconstruction objective}, illustrated in Fig.~\ref{fig:sensoformer}(a). For each training instance, we sample a corruption pattern from a mask bank and hide the corresponding subset of sensor measurements. The mask bank includes point-wise, time-wise, channel-wise, and block-wise masks, which emulate common failure modes in accelerator telemetry such as isolated dropouts, missing time intervals, unavailable sensor channels, and structured channel--time gaps. The visible entries are additionally perturbed with additive noise sampled from multiple noise strengths, including zero, while the mask indicator specifies which entries have been hidden. Given a corrupted input batch and the corresponding binary mask, SensOFormer is trained to reconstruct the original sensor values at the masked locations. By combining structured masking with denoising, the pretraining task discourages simple copying of clean observations and instead encourages the model to learn robust temporal and cross-channel structure from noisy, incomplete accelerator data.

\subsection{Overview of SensOFormer}

Fig.~\ref{fig:sensoformer}(b) illustrates SensOFormer as an encoder--processor--decoder architecture for accelerator sensor sets. The key design choice is to represent the sensors observed at each time step as an unordered set rather than as a fixed channel vector. This allows the model to handle variable sensor availability, channel ordering, and subsystem-specific measurements without tying the representation to a particular global sensor layout. Given a corrupted input batch and its mask indicator, SensOFormer forms per-channel features $x \in \mathbb{R}^{B \times L \times C \times F}$, where $B$ is the batch size, $L$ is the sequence length, $C$ is the number of sensor channels, and $F$ is the feature dimension. In our implementation, $F=2$, corresponding to the corrupted sensor value and a binary mask indicator. The batch and time dimensions are flattened to obtain $(BL,C,F)$, so that each time step is treated as a set of $C$ channel observations. A shared tokenization layer maps these features to channel tokens in $\mathbb{R}^{BL \times C \times d}$.

The \emph{set encoder} uses a Perceiver-style cross-attention mechanism in which $S$ learnable latent queries attend to the channel tokens. Because the same attention operation is applied to an unordered set of sensor tokens, the encoder is permutation-invariant with respect to channel ordering: reordering the input sensors does not change the aggregated latent representation. The resulting slot representations are pooled and reshaped into a latent sequence $z \in \mathbb{R}^{B \times L \times d}$, which is processed by a transformer to model temporal dependencies in a compact shared latent space.

The \emph{set decoder} maps the processed latent sequence back to channel-wise predictions. For each latent vector $z_{b,t}$, the decoder constructs $M$ memory slots, and the channel tokens are reused as output queries that cross-attend to these memory slots. This produces one prediction per queried channel, followed by layer normalization and scalar projection to obtain $\hat{x} \in \mathbb{R}^{B \times L \times C}$. The decoder is permutation-equivariant: reordering the output channel queries reorders the corresponding predictions in the same way. Thus, \emph{SensOFormer combines permutation-invariant encoding, temporal latent processing, and permutation-equivariant decoding to model variable sensor sets without requiring a fixed channel layout}.

\subsection{Downstream tasks}

Fig.~\ref{fig:sensoformer}(c) summarizes how SensOFormer is adapted to downstream tasks after pretraining. We consider three task instantiations: missing-data imputation, anomaly detection, and fault identification. In all cases, the set encoder $E$ and transformer processor $P$ provide a shared representation of the operational sensor data, while the task-specific decoder or head $D_i$ determines the output type. This design allows the same pretrained backbone to support reconstruction, binary classification, and multiclass classification tasks.

For \emph{missing-data imputation}, the input is partially observed telemetry in which missingness can occur as missing sensor values, consecutive missing time steps, missing channel values over multiple time steps, or structured channel–time blocks. Simple interpolation or local averaging can be effective for isolated missing points, but becomes unreliable when multiple time windows or sensor channels are absent. As shown in Fig.~\ref{fig:sensoformer}(c.1), SensOFormer uses the reconstruction decoder $D_1$ to predict the missing values from the available temporal and cross-channel context. This task is directly aligned with the denoising masked reconstruction objective used during pretraining, so the architecture can be reused for missing data reconstruction.

For \emph{anomaly detection}, the goal is to distinguish normal operation from abnormal or faulty machine behavior using operational sensor data. This task is challenging because labeled anomalies are limited: fault annotation requires domain expertise and careful post-hoc analysis. Reliable anomaly detection is operationally important because accelerator facilities run under scheduled beam-time periods and can be interrupted by planned or unplanned maintenance. As shown in Fig.~\ref{fig:sensoformer}(c.2), we adapt the pretrained encoder--processor backbone with a binary classification head $D_2$ to identify normal versus faulty operating patterns. 

For \emph{cavity and fault identification}, the objective is to identify the faulty component or classify the type of fault once faulty data are available. This is a more detailed multiclass classification problem than binary anomaly detection, requiring the model to distinguish between different fault categories or faulty components. As shown in Fig.~\ref{fig:sensoformer}(c.3), we reuse the same pretrained encoder $E$ and processor $P$, and attach a task-specific multiclass head $D_3$ to predict the cavity or fault label, respectively.

\section{Experiments} \label{sec:experiments}

\subsection{Datasets}
The datasets used in this study span diverse accelerator facilities, sensor modalities, and operational tasks. They include multi-subsystem operational measurements from LANSCE at LANL \citep{scheinker2025physics}, beam-loss measurements from LHC at CERN \citep{krymova2023data}, high-voltage converter modulator (HVCM) waveforms for power-electronics fault prognosis at SNS at ORNL \citep{radaideh2023early,radaideh2023operational}, superconducting radio-frequency (SRF) cavity fault data from CEBAF at JLab \citep{tennant2020superconducting}, RF-station anomaly data from LCLS at SLAC \citep{humble2022beam,liang2025coincident}, and GMPS control data from Booster at Fermilab \citep{st2021real}. This collection provides a heterogeneous benchmark for testing whether a single self-supervised sensor model can generalize across facilities, subsystems, sampling structures, and downstream tasks. 

\begin{table}[t]
\centering
\small
\caption{Datasets used for pretraining and downstream evaluation. The benchmark spans operational telemetry and subsystem-specific diagnostics from multiple accelerator facilities. Tensor shapes are reported as $(N,T,C)$, where $N$ is the number of examples, $T$ is the number of time steps, and $C$ is the number of sensors/channels. For SRF-CEBAF, the 192 entries represent extracted features (8 cavities × 4 signals × 6 features per signal), rather than temporal steps. MDI denotes missing-data imputation, and $N^\ast$ denotes a variable number of samples and $C^*$ denotes a variable number of channels, with the bracketed range indicating the minimum and maximum number of channels across the dataset. LANSCE-2025 and GMPS-Booster-Fermilab are evaluation-only, not included in pretraining.}
\label{tab:datasets}
\resizebox{\linewidth}{!}{
\begin{tabular}{llll}
\toprule
Dataset & Facility / subsystem & Dataset/Tensor shape & Downstream task \\
 & & (N,T,C) &  \\
\midrule
LANSCE (2019--2024) & LANSCE operational data & (4250, 10000, $C^* \in [1, 1754]$) & MDI on 10\% test set \\
\midrule
LHC-BL (2017) & \multirow{2}{*}{LHC beam-loss monitoring} & (1, 542061, 66) & \multirow{2}{*}{MDI on 10\% test set} \\
LHC-BL (2018) & & (1, 524430, 66) & \\
\midrule
\multirow{2}{*}{RFS-LCLS (2020)} & LCLS RF station signals (AMM and beam) & (1269, 2196, 9) & \multirow{2}{*}{MDI on 10\% test set} \\
& LCLS RF station signals (AMPL and beam) & (3208, 2076, 9) & \\
\midrule
RFS-LCLS (2024) & LCLS RF station signals (phase and beam) & (5364, 1066, 9) & MDI on 10\% test set \\
\midrule
\multirow{2}{*}{HVCM-SNS} & \multirow{2}{*}{SNS high-voltage converter modulators} & (10000, 3753, 12) & 1. MDI on 10\% test set \\
& & ($N^\ast$, 3753, 12) & 2. Anomaly detection \\
\midrule
\multirow{3}{*}{SRF-CEBAF} & \multirow{3}{*}{CEBAF superconducting RF cavities} & \multirow{3}{*}{(2375, 192, 1)} & 1. MDI on 10\% test set \\
 & &  & 2. Cavity identification \\
& &  & 3. Fault identification \\
\midrule
\midrule
LANSCE (2025) & LANSCE operational data & (17, 10000, $C^* \in [1, 769]$) & MDI on 100\% test set \\ 
\midrule
GMPS-Booster-Fermilab & Gradient Magnet Power Supply & (1,1295713, 43) & MDI on 100\% test set\\
\bottomrule
\vspace{-8mm}
\end{tabular}}
\end{table}

Table~\ref{tab:datasets} summarizes the tensorized form of each dataset and the downstream tasks used in our evaluation. The datasets differ substantially in both temporal length and channel structure: LANSCE contains variable-channel operational telemetry, LHC-BL provides very long beam-loss time series, GMPS-Booster-Fermilab provides long-duration, multivariate operational telemetry capturing cycle-by-cycle readings and control settings, RFS-LCLS and HVCM-SNS contain fixed-channel waveform-style records, and SRF-CEBAF contains feature vectors extracted from RF fault events. All datasets are used for missing-data imputation (MDI), while HVCM-SNS and SRF-CEBAF additionally support diagnostic tasks such as anomaly detection, cavity identification, and fault identification. This design allows us to evaluate both zero-shot reconstruction across heterogeneous sensor formats and transfer to task-specific accelerator diagnostics.

\subsection{Main results}
For readability and conciseness, we defer details of the experimental settings, including model variants, windowing, masking, normalization, noise augmentation, the training objective, optimization, and compute, to Appendix~\ref{app:experimental_settings}.

A detailed \emph{scaling study} in Appendix~\ref{app:scaling_studies} shows that validation masked-reconstruction loss decreases with both model size and pretraining data volume, with the best fitted point occurring at the largest evaluated model--data scale. In our setting, the fitted exponents are $\alpha=0.715$ for model size and $\beta=0.374$ for data size, indicating that both larger models and larger pretraining corpora improve reconstruction loss within the evaluated regime. The resulting compute-optimal trend allocates additional compute to both dimensions, with data size increasing faster than parameter count, suggesting that future accelerator foundation models should scale the pretraining corpus alongside model capacity.

The main text focuses on three empirical questions that directly test whether SensOFormer learns reusable accelerator representations: (i) can the pretrained model perform zero-shot missing-data imputation across heterogeneous datasets spanning six accelerator facilities? (ii) can the frozen pretrained backbone support early anomaly detection on HVCM-SNS with only a lightweight supervised head? and (iii) can the same representation transfer to cavity and fault identification on SRF-CEBAF? We report these results in the following subsections.

\subsubsection{Zero-shot missing-data imputation}
Table~\ref{tab:zeroshot_pointwise} reports zero-shot point-wise imputation performance across datasets, mask ratios, and model variants: XTi (2.38M), Ti (5.80M), Mi (13.12M), S (35.87M), and B (80.54M). Lower values indicate better reconstruction. Overall, the B model achieves the lowest error in 25 of the 28 dataset--mask combinations, while the S model is best in the remaining three settings. This shows a clear benefit from increasing model capacity for zero-shot imputation, suggesting that larger SensOFormer variants more effectively use temporal and cross-channel context to reconstruct missing sensor values. 

The \emph{LANSCE-2025} setting is more challenging than the LANSCE 2019--2024 held-out test set because it evaluates transfer to a different operating year rather than to a held-out test set from the same pretraining-year distribution. The reconstruction errors are substantially higher on LANSCE-2025, indicating a temporal distribution shift in the sensor statistics, operating conditions, or available channel coverage. Representative point-wise and block-wise reconstructions on the unseen year are shown in Figs.~\ref{fig:viz_mdi_point_ds1} and \ref{fig:viz_mdi_block_ds1}.

To further assess cross-facility generalization, we evaluate \emph{GMPS-Booster-Fermilab}, a previously unseen accelerator facility excluded entirely from pretraining. Without any facility-specific fine-tuning, SensOFormer achieves zero-shot imputation on GMPS-Booster, with the B variant obtaining the lowest reconstruction error across all four masking ratios. At 15\% masking, the MAE decreases from 154.46 for XTi to 148.07 for B. These results extend our evaluation to demonstrate zero-shot transfer to an unseen accelerator facility with a different sensor configuration. Representative reconstructions on the unseen GMPS-Booster-Fermilab dataset are shown in Fig.~\ref{fig:viz_mdi_gmps}.

\begin{table}[t]
\centering
\small
\caption{Zero-shot imputation performance on held-out test sets, reported as raw (un-normalized scale) mean absolute error (MAE) averaged over masked entries for all the samples in the test set. \emph{Lower values indicate better reconstruction}. Bold values denote the best model for each dataset and mask ratio.}
\label{tab:zeroshot_pointwise}
\resizebox{0.85\linewidth}{!}{%
\begin{tabular}{llccccc}
\toprule
Dataset & Mask ratio & XTi & Ti & Mi & S & B \\
\midrule
\multirow{4}{*}{LANSCE 2019--2024}
& 5\%  & 2.4928 & 1.9969 & 2.0296 & 1.8414 & \textbf{1.7130} \\
& 15\% & 2.4693 & 1.9984 & 1.9526 & 1.8916 & \textbf{1.6769} \\
& 25\% & 2.5411 & 2.1002 & 1.9933 & 2.0337 & \textbf{1.7580} \\
& 50\% & 2.6005 & 2.3238 & 2.1687 & 2.1835 & \textbf{2.0624} \\
\midrule
\multirow{4}{*}{LHC-BL}
& 5\%  & 3.0867 & 2.8527 & 2.6608 & 2.4952 & \textbf{2.3110} \\
& 15\% & 3.1752 & 2.9858 & 2.8162 & 2.5671 & \textbf{2.5179} \\
& 25\% & 3.1206 & 3.1030 & 2.9639 & \textbf{2.5627} & 2.5879 \\
& 50\% & 2.9752 & 2.9677 & 3.0199 & 2.8093 & \textbf{2.7347} \\
\midrule
\multirow{4}{*}{HVCM-SNS}
& 5\%  & 136.5608 & 166.7894 & 134.7735 & 130.5587 & \textbf{127.9945} \\
& 15\% & 160.5342 & 157.6206 & 151.5387 & \textbf{139.9381} & 140.3383 \\
& 25\% & 183.9417 & 189.1077 & 177.0788 & 170.1272 & \textbf{167.9301} \\
& 50\% & 201.4619 & 219.4875 & 198.7487 & 207.8971 & \textbf{197.6245} \\
\midrule
\multirow{4}{*}{RFS-LCLS}
& 5\%  & 0.9644 & 0.8713 & 0.8095 & 0.7624 & \textbf{0.7234} \\
& 15\% & 0.9726 & 0.8775 & 0.8295 & 0.7489 & \textbf{0.7220} \\
& 25\% & 1.0196 & 0.9111 & 0.8763 & 0.8017 & \textbf{0.7630} \\
& 50\% & 1.1120 & 1.0106 & 1.0497 & 0.8997 & \textbf{0.8896} \\
\midrule
\multirow{4}{*}{SRF-CEBAF}
& 5\%  & 0.2306 & 0.2288 & 0.2347 & 0.2177 & \textbf{0.2099} \\
& 15\% & 0.2268 & 0.2278 & 0.2264 & 0.2197 & \textbf{0.2123} \\
& 25\% & 0.2241 & 0.2220 & 0.2228 & 0.2194 & \textbf{0.2140} \\
& 50\% & 0.2247 & 0.2182 & 0.2248 & 0.2208 & \textbf{0.2137} \\
\midrule
\midrule
\multirow{4}{*}{LANSCE-2025}
& 5\%  & 16.4059 & 11.0818 & 6.7363 & 6.8370 & \textbf{6.1174} \\
& 15\% & 8.4765 & 7.2481 & 7.7511 & 7.3025 & \textbf{6.6987} \\
& 25\% & 7.9151 & 7.3074 & 8.1885 & 7.8608 & \textbf{6.9565} \\
& 50\% & 8.4266 & 8.1807 & 7.9797 & \textbf{7.4946} & 8.6389 \\
\midrule
\multirow{4}{*}{GMPS-Booster}
& 5\%  & 154.4794 & 152.1034 & 152.9284 & 151.7627 & \textbf{148.5698} \\
& 15\% & 154.4622 & 150.8483 & 150.4868 & 148.5588 & \textbf{148.0681} \\
& 25\% & 156.9602 & 154.9191 & 154.6353 & 150.1825 & \textbf{150.0526} \\
& 50\% & 158.7238 & 156.6129 & 156.7293 & 155.2808 & \textbf{154.2139} \\
\bottomrule
\vspace{-8mm}
\end{tabular}}
\end{table}

Qualitative reconstructions in Figs.~\ref{fig:viz_mdi_point_ds0}-\ref{fig:viz_mdi_gmps} (Appendix~\ref{app:extended_results}) further support the quantitative results of Table~\ref{tab:zeroshot_pointwise}. The predicted values at masked locations closely follow the normalized ground-truth trajectories in representative held-out windows, showing that the model is not only reducing aggregate MAE but also recovering local temporal structure. Together, the quantitative and qualitative results demonstrate that masked-denoising pretraining learns reusable sensor representations that can be applied directly to missing-data imputation across heterogeneous accelerator datasets.

\subsubsection{Anomaly detection task}

We formulate supervised anomaly detection as a binary classification problem on the HVCM-SNS dataset. The labeled dataset contains 10,000 waveform samples, split into 8,000 training samples and 2,000 held-out test samples, and the operationally relevant evaluation is performed on 21 induced early-fault test files following the protocol of \citet{radaideh2023early}. Each input consists of multichannel HVCM waveforms, and the model predicts whether the signal corresponds to normal operation or a faulty condition.

To adapt SensOFormer to this task, we freeze the pretrained set encoder and latent transformer processor and train only a lightweight binary classification head on top of the latent sequence representation. The head is a one-dimensional temporal CNN that operates on the latent sequence $z \in \mathbb{R}^{B \times L\times d}$ with four Conv1D blocks with GELU activations and max pooling after each convolution. The resulting temporal features are summarized using both global max pooling and global mean pooling, concatenated, normalized, and passed through a three-layer MLP before producing a single binary logit. The classification head of XTi contains 0.22M trainable parameters.

We train the detector head for 200 epochs using AdamW with a learning rate of $10^{-5}$. On the 8,000/2,000 train/test split, all methods, including SensOFormer, achieve perfect precision, recall, F1, and AUC, making this split insufficient for distinguishing practical early-warning performance. As shown in Table~\ref{tab:sns2_results}, SensOFormer-XTi passes 20 of the 21 induced early-fault tests, matching the strongest prior hierarchical voting ensemble. Scaling to SensOFormer-B improves this result to 21 of 21 tests (100\% success rate), outperforming both XTi and the strongest prior method. Since both pretrained backbones are frozen and use the same classification head, this improvement provides evidence that increasing backbone capacity improves downstream diagnostics.

\begin{table}[h]
\vspace{-3mm}
\centering
\small
\caption{All 11 evaluated models, including nine prior models and SensOFormer-XTi and B, achieved perfect precision, recall, F1, and AUC on the 8000/2000 train/test split. The main comparison is the 21 induced early-fault tests, where a test is counted as passed when $\rho < 25\%$ and F1 $> 0.8$, following \cite{radaideh2023early}.}
\label{tab:sns2_results}
\begin{tabular}{lcc}
\toprule
Model & Passed tests & Success rate (\%) \\
\midrule
Hierarchical voting ensemble (VE) & \textbf{20 / 21} & \textbf{95.2} \\
AdaBoost (AB)                     & 11 / 21 & 52.4 \\
Extremely randomized trees (ET)   & 10 / 21 & 47.6 \\
CNN                               & 9 / 21  & 42.9 \\
Bagging classifier (BC)           &  9 / 21  & 42.9 \\
Random forest (RF)                &  9 / 21  & 42.9 \\
Gradient boosting (GB)            & 9 / 21  & 42.9 \\
Linear SVM (LSVM)                 & 5 / 21  & 23.8 \\
RBF-SVM                           & 5 / 21  & 23.8 \\
\textbf{SensOFormer-XTi (Ours)}   & \textbf{20 / 21} & \textbf{95.2} \\
\textbf{SensOFormer-B (Ours)}     & \textbf{21 / 21} & \textbf{100} \\
\bottomrule
\vspace{-7mm}
\end{tabular}
\end{table}

\begin{table}[h]
\centering
\small
\caption{Test-set accuracy for CEBAF C100 cavity identification and fault identification. We compare SensOFormer-XTi with prior deep-learning baselines from \citet{shabalina2020machine}. SensOFormer-XTi matches the strongest baselines on cavity identification and improves fault identification accuracy from 74.10\% to 84.01\%.}
\label{tab:cebaf_fault_cavity}
\begin{tabular}{lcccc}
\toprule
Task & LSTM & CNN, direct mapping & CNN, scalogram & \textbf{SensOFormer-XTi (Ours)} \\
\midrule
Cavity identification & 84.60 & 84.60 & 45.00 & \textbf{84.80} \\
Fault identification  & 74.10 & 74.10 & 58.90 & \textbf{84.01} \\
\bottomrule
\vspace{-6mm}
\end{tabular}
\end{table}

\subsubsection{Fault and cavity identification task}
We evaluate SensOFormer on supervised SRF fault diagnosis using the SRF-CEBAF dataset. The CEBAF C100 cryomodule data support two related multiclass classification tasks: \emph{cavity identification}, where the goal is to identify the first faulting cavity, and \emph{fault identification}, where the goal is to classify the fault type. These tasks are challenging because SRF cavities in a cryomodule are strongly coupled, so a primary fault can quickly induce secondary responses in neighboring cavities.

To adapt SensOFormer to these tasks, we reuse the pretrained encoder--processor backbone and replace the reconstruction decoder with a lightweight temporal CNN classification head. For each event, the backbone produces a latent sequence $z \in \mathbb{R}^{B \times 192 \times d}$, where $192$ corresponds to the extracted feature dimension and $d$ is the latent feature dimension. We train separate heads for cavity identification and fault identification, with output dimensions of 9 and 8 classes, respectively. The head first applies layer normalization to the latent sequence, then processes it with two one-dimensional convolutional layers along the 192 feature-index dimension, followed by three fully connected layers. This task-specific head contains approximately $0.5$M trainable parameters. We train for 200 epochs using AdamW with a learning rate of $10^{-4}$ and batch size of 32.

Table~\ref{tab:cebaf_fault_cavity} compares SensOFormer-XTi with prior deep-learning baselines on the CEBAF test set. For cavity identification, SensOFormer-XTi achieves 84.80\% accuracy, comparable to the best prior LSTM and CNN direct-mapping baselines. For fault identification, SensOFormer-XTi achieves 84.01\% accuracy, substantially improving over the prior 74.10\% baselines. These results indicate that the pretrained SensOFormer representation is useful not only for reconstruction, but also for supervised diagnostic tasks that require distinguishing short RF waveform fault signatures.

\begin{wrapfigure}{r}{0.48\textwidth}
\vspace{-1.0em}
\centering
\includegraphics[width=0.43\textwidth, trim=0cm 0cm 0cm 0cm, clip]{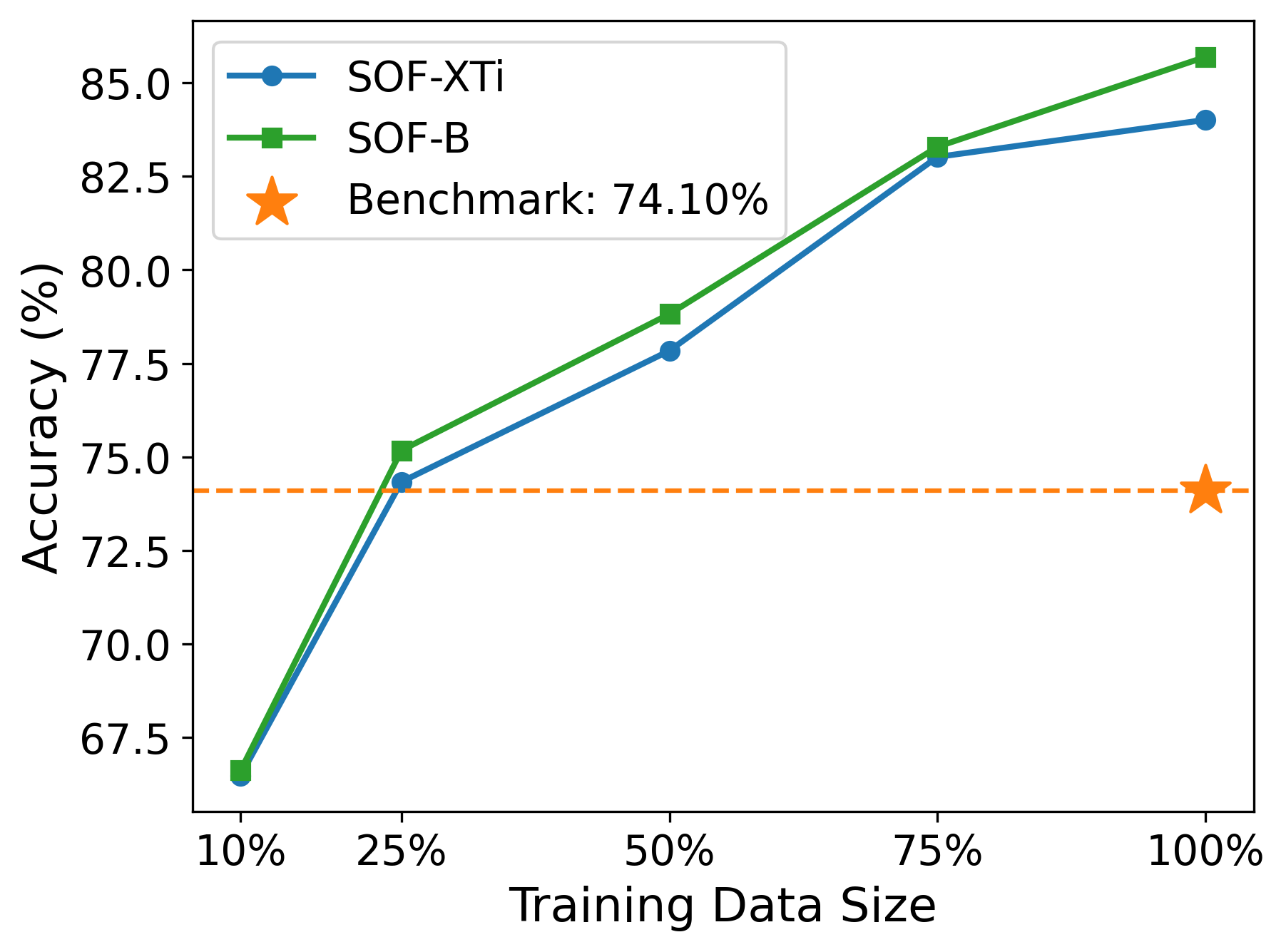}
\captionsetup{justification=centering}
\caption{Fault-identification accuracy.}
\label{fig:cebaf_datascarce}
\vspace{-1.0em}
\end{wrapfigure}

We further evaluate fault identification in a data-scarce setting, motivated by the fact that accelerator fault data are costly to collect and require expert annotation. To assess label efficiency, we vary the fraction of labeled training examples while keeping the test set fixed. As shown in Fig.~\ref{fig:cebaf_datascarce}, SensOFormer-XTi improves monotonically as additional labeled data are introduced and surpasses the prior deep-learning benchmark using only 25\% of the training set. 

Scaling the model from XTi (2.38M parameters) to B (80.54M parameters) consistently improves accuracy across all training fractions, reaching 85.69\% compared with 84.01\% at full training data. Both models use frozen pretrained backbones with the same classification-head architecture, labeled-data splits, and optimization settings. These results suggest that self-supervised pretraining provides useful representations for label-efficient SRF fault diagnosis, with additional improvements from increased model capacity. Further details of the published diagnostic baseline architectures, evaluation protocols, and parameter-count estimates are provided in Appendix~\ref{app:experimental_settings}.

\paragraph{Limitations of current research.}
Our evaluation covers multiple accelerator facilities and subsystems, but remains limited to the available public and archived datasets. SensOFormer incurs substantial one-time pretraining costs, while larger backbones increase inference costs. Task-specific models may remain preferable for fixed sensor configurations and narrowly defined tasks. For substantial distribution shifts, parameter-efficient fine-tuning may provide a practical adaptation strategy. The present study does not investigate continual learning, which remains an important direction for adapting to changing accelerator operating conditions. In addition, SensOFormer is evaluated offline. Real-time deployment and integration with accelerator control systems remain future work.

\section{Conclusion}

We introduced SensOFormer, a self-supervised foundation model for particle-accelerator operational data. The design is well suited to accelerator telemetry, where sensors can vary across facilities, subsystems, acquisition windows, and operating modes. By pretraining with masked denoising reconstruction, SensOFormer learns temporal and cross-channel structure from noisy, incomplete, and diverse accelerator measurements. Across datasets from LANSCE, LHC, SNS, LCLS, CEBAF, and Booster, SensOFormer demonstrates strong transfer across missing-data reconstruction, anomaly detection, cavity identification, and fault identification. The scaling study further shows that reconstruction performance improves with both model size and pretraining data, suggesting that accelerator foundation models can benefit from systematic scaling. These results support the feasibility of a general-purpose sensor foundation model for large-scale scientific facilities. Instead of training separate models for each subsystem or diagnostic task, SensOFormer provides a reusable backbone that can be adapted to multiple accelerator problems with limited additional supervision. 

\section*{Data and code availability}
The datasets used in this work are cited in the paper. Publicly available datasets are accessed through their original releases. Any datasets subject to facility-specific access restrictions are not redistributed. Code along with explainer video is available on the project page: \url{https://mahindrautela.github.io/sensoformer}.

\section*{Acknowledgements}
Research presented in this article was supported by the Programmatic AI for LANSCE effort
(internal charge code: CFLC) through the AOT-IC group at LANL. This research used resources provided by the LANL Institutional Computing Program, which is supported by the U.S. DOE/NNSA under Contract No. 89233218CNA000001.

\bibliographystyle{abbrvnat}
\bibliography{references}

\appendix

\newpage

\section{Extended Related Work} \label{app:related_works}

\paragraph{Relation to set-based and any-variate time-series models.}
Set-based processing of irregular observations, arbitrary-variate modeling,
and multi-source self-supervised pretraining are established ideas in
time-series learning. SeFT~\citep{horn2020set} represents irregularly sampled
multivariate time series as an unordered set of observation events and
aggregates them with set functions, while MOIRAI~\citep{woo2024unified}
employs any-variate attention to accommodate an arbitrary number of variates
and is pretrained on a large multi-domain corpus for zero-shot forecasting.
SensOFormer differs from these models across several architectural, training,
and task dimensions, as summarized in Table~\ref{tab:seft_moirai_comparison}.
Our contribution therefore does not lie in any one of these components in
isolation. Rather, it lies in integrating latent channel-set compression,
temporal representation learning, permutation-equivariant reconstruction,
structured telemetry corruption, and multi-facility pretraining for
accelerator imputation and diagnostic tasks.

\paragraph{Applicability of MOIRAI to accelerator telemetry.}
We also examined the released Uni2TS implementation of
MOIRAI\footnote{\url{https://github.com/SalesforceAIResearch/uni2ts}} for
zero-shot inference and fine-tuning on accelerator datasets. The released
formulation is not directly task-equivalent to our benchmark. Its default
fine-tuning configuration sets the maximum number of variates to 128, whereas
our datasets contain $C \in [1, 1754]$ channels. A comparable adaptation would
therefore require changes to the masking pipeline, output heads, losses, and
high-dimensional channel handling, as summarized in
Table~\ref{tab:moirai_applicability}. More broadly, accelerator telemetry
combines facility-dependent sensor layouts, up to 1,754 channels,
heterogeneous sampling, structured missingness and noise, and limited
diagnostic labels; SeFT and MOIRAI are therefore not directly applicable
without changing their input representation, training objective, and output
modules.

\begin{table}[h]
\centering
\caption{Comparison of SensOFormer with the set-based irregular time-series
model SeFT~\citep{horn2020set} and the any-variate time-series foundation
model MOIRAI~\citep{woo2024unified}.}
\label{tab:seft_moirai_comparison}
\footnotesize
\renewcommand{\arraystretch}{1.15}
\begin{tabularx}{\linewidth}{@{}>{\raggedright\arraybackslash}p{2.1cm}
    >{\raggedright\arraybackslash}X
    >{\raggedright\arraybackslash}X
    >{\raggedright\arraybackslash}X@{}}
\toprule
 & \textbf{SeFT} & \textbf{MOIRAI} & \textbf{SensOFormer (Ours)} \\
\midrule
Input token
  & Observation event $(t, z, m)$
  & Time--variate patch
  & Sensor-channel set at each time step \\
Channel encoder
  & Independent MLPs with Deep-Set aggregation
  & Any-variate full self-attention
  & Perceiver-style latent set encoder with cross-attention \\
Temporal modeling
  & No sequential processor; time is encoded independently for each event
  & Joint time--variate encoder Transformer
  & Separate latent temporal Transformer \\
Decoder / output
  & Attention pooling followed by a classification MLP
  & Linear patch projection to probabilistic mixture parameters
  & Permutation-equivariant set decoder with cross-attention \\
Self-supervised pretraining 
  & \xmark & \cmark & \cmark \\
Training objective
  & Supervised classification
  & Self-supervised probabilistic forecasting
  & Self-supervised denoising masked reconstruction \\
Masking / corruption
  & \xmark
  & Future-horizon patch masking
  & Point, time, channel, and block masking with noise corruption \\
Training data
  & Three clinical datasets
  & LOTSA: 27 billion observations across nine domains
  & Five particle-accelerator facilities \\
Data characteristics
  & Irregular and asynchronous measurements with approximately 37--40 modalities
  & Multiple regular sampling frequencies with up to 128 variates
  & Tensorized heterogeneous data with 1--1754 sensor channels \\
Scales to $>10^3$ channels 
  & \xmark & \xmark & \cmark \\
Downstream tasks
  & Mortality and sepsis classification
  & Zero-shot probabilistic forecasting
  & Imputation and diagnostic classification \\
\bottomrule
\end{tabularx}
\end{table}

\begin{table}[t]
\centering
\caption{Differences between the released MOIRAI (Uni2TS) formulation and the
requirements of our accelerator benchmark.}
\label{tab:moirai_applicability}
\footnotesize
\renewcommand{\arraystretch}{1.15}
\begin{tabularx}{\linewidth}{@{}>{\raggedright\arraybackslash}p{2.0cm}
    >{\raggedright\arraybackslash}X
    >{\raggedright\arraybackslash}X@{}}
\toprule
 & \textbf{Released MOIRAI (Uni2TS)} & \textbf{Accelerator benchmark requirement} \\
\midrule
Task
  & Predicts a future time-series horizon.
  & Reconstructs missing values occurring anywhere in the sensor sequence. \\
Masking
  & Masks complete future patches.
  & Uses point-wise, time-wise, channel-wise, and block-wise masks. \\
Mask granularity
  & Prediction masks operate at the patch-token level.
  & Missingness is defined for individual sensor values. \\
Output head
  & Uses a probabilistic forecasting output head.
  & Requires anomaly, cavity, and fault-classification heads. \\
Number of variates
  & The default fine-tuning configuration supports at most 128 variates.
  & Accelerator samples contain as many as 1,754 sensor channels. \\
Attention
  & Flattens time--variate patches and applies full self-attention.
  & Requires efficient processing of very large and variable sensor sets. \\
Input structure
  & Fine-tuning examples use fixed-length, fixed-shape sliding windows.
  & Channel counts and sequence structures differ across facilities. \\
Overall scope
  & Natively supports probabilistic future-horizon forecasting.
  & Our benchmark requires arbitrary-gap imputation and diagnostic
    classification over heterogeneous datasets with up to 1,754 channels. \\
\bottomrule
\end{tabularx}
\end{table}

\section{Experimental settings} \label{app:experimental_settings}

\paragraph{Model variants.}
We evaluate five SensOFormer variants, summarized in Table~\ref{tab:sensoformer_variants}. The variants differ in the latent attention dimension, number of attention heads, and transformer depth, while keeping the same encoder--processor--decoder architecture. For all variants, the MLP dimension is set to $4d$, where $d$ is the attention dimension. Unless otherwise specified, the set encoder uses 16 latent slots, the set decoder uses 32 memory slots, gated pooling is used in the set encoder, and the transformer uses absolute positional encodings with maximum length 1024. We restrict the largest model to the B-scale variant based on the scaling study in Appendix~\ref{app:scaling_studies}, where the performance gains begin to show diminishing returns relative to the additional training cost.

\begin{table}[h]
\centering
\small
\caption{SensOFormer model variants used in this study. The MLP dimension is set
to $4d$ for all variants, where $d$ is the attention dimension.}
\label{tab:sensoformer_variants}
\begin{tabular}{lccccc}
\toprule
Model & Attention dim. $d$ & Heads & Depth & MLP dim. & Parameters \\
\midrule
XTi & 192 & 4  & 2 & 768  & 2.38M \\
Ti  & 256 & 4  & 4 & 1024 & 5.80M \\
Mi  & 384 & 6  & 4 & 1536 & 13.12M \\
S   & 512 & 8  & 8 & 2048 & 35.87M \\
B   & 768 & 12 & 8 & 3072 & 80.54M \\
\bottomrule
\end{tabular}
\end{table}

\paragraph{Pretraining data and windowing.}
SensOFormer is pretrained on the heterogeneous accelerator datasets described in Table~\ref{tab:datasets}, including LANSCE, LHC-BL, HVCM-SNS, RFS-LCLS, and SRF-CEBAF data. Each original time series is treated as a separate sequence, and the train/validation split is performed at the sequence level before windowing. This prevents overlapping windows from the same original sequence from appearing in both training and validation sets. Unless otherwise specified, 80\% of the available sequences are used for pretraining and 10\% each are used for validation and test. Each sequence is then segmented into overlapping windows of length $L=1024$ with stride 256. The number of sensor channels is kept dataset-specific, and the model is trained without padding along the channel dimension. This is possible because the set encoder and set decoder operate on sensor sets rather than on a fixed global channel layout.

\paragraph{Masking.}
For each training window, we uniformly sample one mask type from a mask bank and train the model to reconstruct the held-out entries from the remaining visible measurements. The mask bank contains four masking patterns: point-wise, time-wise, channel-wise, and block-wise masking. Point-wise masking independently masks entries with probability $0.15$. Time-wise masking samples one time index and masks a fraction $0.75$ of the channels at that time index. Channel-wise masking samples one sensor channel and masks a fraction $0.60$ of its time steps. Block-wise masking samples a contiguous rectangular region in the
time--channel plane with area approximately $0.15LC$, where $L$ is the window length and $C$ is the number of channels; the block aspect ratio is sampled randomly. For the time-wise, channel-wise, and block-wise masks, we avoid masking
an entire row, column, or window whenever possible. These patterns emulate common missingness modes in accelerator telemetry, including isolated dropouts, partial time-slice loss, intermittent sensor-channel loss, and structured channel--time
gaps. Masked entries are replaced by a sentinel value, while the binary mask is provided as an input feature so that the model can distinguish intentionally hidden entries from valid observed measurements.

\paragraph{Normalization.}
Because the pretraining corpus combines sensors with different physical units, dynamic ranges, and operating regimes, we normalize each window on the fly rather than using global dataset statistics. Normalization is performed independently
for each batch element and sensor channel, using only the visible entries in that window. Let $x_{b,t,c}$ denote the value for batch element $b$, time index $t$, and channel $c$, and let $M_{b,t,c}=1$ indicate a masked target. For each $(b,c)$ pair, we compute
\begin{equation}
x^{\min}_{b,c}=\min_{t:M_{b,t,c}=0} x_{b,t,c},
\qquad
x^{\max}_{b,c}=\max_{t:M_{b,t,c}=0} x_{b,t,c},
\end{equation}
and define the normalization range as
\begin{equation}
r_{b,c}=\max\left(x^{\max}_{b,c}-x^{\min}_{b,c},\epsilon\right),
\end{equation}
with $\epsilon=10^{-5}$. The normalized signal is then
\begin{equation}
\tilde{x}_{b,t,c}
=
\mathrm{clip}\left(
\frac{x_{b,t,c}-x^{\min}_{b,c}}{r_{b,c}},
0,1
\right).
\end{equation}
Computing the statistics only from unmasked entries prevents information from the held-out targets from leaking into the input normalization. This per-window, per-channel normalization removes dataset-specific scale differences while preserving the relative temporal structure within each sensor. The same masked normalization procedure is used during validation and downstream imputation.

\paragraph{Noise augmentation.}
After normalization, we apply denoising augmentation during pretraining. For each training batch, we sample a noise level
\begin{equation}
\sigma \in \{0,\;0.005,\;0.01,\;0.015\},
\end{equation}
and add zero-mean Gaussian noise only to the visible entries:
\begin{equation}
\tilde{x}^{\mathrm{in}}_{b,t,c}
=
\tilde{x}_{b,t,c}
+
(1-M_{b,t,c})\epsilon_{b,t,c},
\qquad
\epsilon_{b,t,c}\sim\mathcal{N}(0,\sigma^2).
\end{equation}
Masked entries are not perturbed by this noise process; they are represented to the model through the mask indicator and sentinel value. Including $\sigma=0$ keeps a subset of batches noise-free, while the nonzero noise levels encourage
the model to learn robust denoising representations instead of simply copying the visible measurements. Noise augmentation is used only during training and is not applied during validation or zero-shot imputation.

\paragraph{Training objective.}
The model is trained to reconstruct the clean normalized signal from the corrupted input. Let $x$ be the normalized target, $\hat{x}$ be the model prediction, and $M$ be the binary mask, where $M_{t,c}=1$ indicates a masked entry. The training loss is

\begin{equation}
\begin{aligned}
\mathcal{L}
&=
\frac{\sum_{t,c} M_{t,c}(\hat{x}_{t,c}-x_{t,c})^2}
     {\sum_{t,c} M_{t,c}}
+
\lambda_{\mathrm{unmask}}
\frac{\sum_{t,c} (1-M_{t,c})(\hat{x}_{t,c}-x_{t,c})^2}
     {\sum_{t,c} (1-M_{t,c})}
\\
&\quad+
\lambda_{\mathrm{smooth}}
\frac{1}{(L-1)C}
\sum_{t=1}^{L-1}\sum_{c=1}^{C}
\left|\hat{x}_{t+1,c}-\hat{x}_{t,c}\right|.
\end{aligned}
\end{equation}

The first term is the masked reconstruction loss and is the primary pretraining objective. The second term weakly regularizes reconstruction on unmasked entries, and the third term penalizes high-frequency variation in the predicted sequence. We use $\lambda_{\mathrm{unmask}}=0.05$ and $\lambda_{\mathrm{smooth}}=0.01$. Validation loss is computed only on masked entries, matching the missing-data imputation objective.

\paragraph{Optimization and compute.}
Pretraining is performed using distributed data parallelism on 16 NVIDIA H100 GPUs with 80GB memory each. Each GPU processes one micro-batch, and gradient accumulation is used before applying an optimizer update. With 16 GPUs, a per-GPU micro-batch size of 1, and accumulation over 4 local micro-batches per GPU, the effective global batch size is $16 \times 1 \times 4 = 64$. All SensOFormer variants are pretrained for 25 epochs, where one epoch denotes one pass through the pretraining set. This corresponds to approximately 125K optimizer updates over the full pretraining run. We use the same optimization recipe for all model sizes so that performance differences are primarily attributable to model scale. We use AdamW with a learning rate of $10^{-5}$ and the default PyTorch weight decay. The learning rate follows a warmup-cosine schedule: the first 2.5\% of optimizer updates are used for linear warmup, followed by cosine decay over the remaining updates. We use automatic mixed precision with bfloat16 to reduce memory usage and improve throughput,
while computing the reconstruction loss in full precision for numerical stability. Gradients are clipped to a maximum norm of 1.0 before each optimizer update. All downstream inference and supervised adaptation experiments were performed on a single NVIDIA RTX A6000 GPU with 48 GB of memory.

\paragraph{HVCM-SNS anomaly-detection baselines.}
The HVCM-SNS benchmark contains 10,000 pulses with 12 waveforms and 3,753 time samples per pulse, divided into 8,000 training and 2,000 held-out test samples. Published baselines include convolutional neural networks, tree-based ensembles, support vector machines, and a hierarchical voting ensemble. The published CNN comprises four Conv1D blocks followed by three fully connected
layers and was trained for five epochs using Adam, a learning rate of $5\times 10^{-4}$, and a batch size of 64. The reported tree-based methods use 40 estimators.

The 21 induced early-fault test files are separate experimental runs excluded from training and standard testing. Each run begins with approximately three minutes of normal operation, followed by gradual perturbation of HVCM settings to induce progressive waveform degradation. Following the published evaluation protocol, a test is counted as passed when the normalized detection delay
satisfies $\rho < 25\%$ and $F_1 > 0.8$. On this benchmark, both SensOFormer-XTi and SensOFormer-B use frozen pretrained encoder--processor backbones and the same temporal CNN classification-head architecture, with approximately $0.22$M downstream trainable parameters. The full pretrained
backbone is not optimized during this supervised adaptation.

\paragraph{CEBAF fault-identification baseline parameter estimates.}
The published CEBAF baselines consist of a three-layer bidirectional LSTM, a direct-mapping CNN, and a scalogram-input CNN. We reproduce the parameter estimates from the architectures described in \citep{shabalina2020machine}. Since the final classification-layer dimensions are not fully specified, the calculated values are lower bounds, not exact total network parameter counts.

\emph{Bidirectional LSTM.} The three bidirectional recurrent layers use hidden size $h=64$. The first layer receives $d=136$ input features, whereas each of the following two layers receives $d=128$ features. A bidirectional LSTM layer has
\begin{equation}
P_{\mathrm{BiLSTM}}(d,h) = 2\left(4hd+4h^2+8h\right)
\end{equation}
parameters, including recurrent and input weights and biases. Consequently,
\begin{align}
P_{\mathrm{LSTM}}
&= P_{\mathrm{BiLSTM}}(136,64)
 + 2P_{\mathrm{BiLSTM}}(128,64) \\
&= 302{,}080 \approx 0.302\,\mathrm{M}.
\end{align}

\emph{Direct-mapping CNN.} The reported CNN comprises five convolutional layers, each with 64 output channels, and batch normalization. The first layer has one input channel and the other four layers have 64 input channels. Using the reported $5\times5$ kernels and including convolutional biases and batch-normalization scale and shift parameters gives
\begin{align}
P_{\mathrm{CNN,direct}}
&= (5\times5\times1+1)64 \nonumber\\
&\quad+4(5\times5\times64+1)64 \nonumber\\
&\quad+5(2\times64) \\
&= 412{,}160 \approx 0.412\,\mathrm{M}.
\end{align}

\emph{Scalogram-input CNN.} The scalogram CNN uses the same architecture, except that its first convolution receives 32 input channels. Thus,
\begin{align}
P_{\mathrm{CNN,scalogram}}
&= (5\times5\times32+1)64 \nonumber\\
&\quad+4(5\times5\times64+1)64 \nonumber\\
&\quad+5(2\times64) \\
&= 461{,}760 \approx 0.462\,\mathrm{M}.
\end{align}
The estimates for all three CEBAF baselines exclude any unspecified final
classification layers.

\paragraph{Parameter-count summary and interpretation.}
Table~\ref{tab:diagnostic_param_counts} summarizes the estimated baseline parameter counts alongside the pretrained SensOFormer model sizes and trainable downstream heads. The complete pretrained SensOFormer-XTi and SensOFormer-B models contain $2.38$M and $80.54$M parameters, respectively
(Table 7). Their downstream evaluation uses frozen encoder--processor backbones; the pretraining reconstruction decoder is not part of the classification head. CEBAF classification uses an approximately $0.50$M-parameter temporal CNN head, whereas HVCM anomaly detection uses an
approximately $0.22$M-parameter head. The XTi and B comparisons use the same classification-head designs and training protocols. The exact number of head parameters can depend on the backbone's latent width, so the stated head counts are approximate.

The main controlled comparison is between SensOFormer-XTi and SensOFormer-B: the main change is pretrained backbone scale, while the backbone remains frozen and the classification-head architecture, label fractions, and optimization protocol are held the same. This tests the effect of pretrained representation capacity on downstream discrimination without claiming that the total model parameters match those of the published task-specific baselines. The larger backbone also incurs greater feature-extraction cost, which is not quantified by trainable-head parameter counts alone.

\begin{table}[h]
\centering
\small
\caption{Parameter-count comparison for published CEBAF baselines and
SensOFormer. Baseline counts are lower bounds inferred from the specified
layers; head counts refer only to trainable downstream parameters, while
pretrained-model counts refer to the complete pretrained architecture.
These categories are not parameter-matched totals.}
\label{tab:diagnostic_param_counts}
\begin{tabular}{@{}p{0.40\linewidth}cp{0.34\linewidth}@{}}
\hline
\textbf{Model/component} & \textbf{Parameters} & \textbf{Scope} \\
\hline
CEBAF bidirectional LSTM & $\geq 0.302$M & Specified recurrent layers \\
CEBAF direct-mapping CNN & $\geq 0.412$M & Convolutions and batch norm \\
CEBAF scalogram CNN & $\geq 0.462$M & Convolutions and batch norm \\
SensOFormer-XTi & $2.38$M & Full pretrained model \\
SensOFormer-B & $80.54$M & Full pretrained model \\
SensOFormer HVCM head & $\sim 0.22$M & Trainable head only \\
SensOFormer CEBAF head & $\sim 0.50$M & Trainable head only \\
\hline
\end{tabular}
\end{table}

\newpage
\section{Scaling studies} \label{app:scaling_studies}

Scaling studies are important in the foundation-model setting because they quantify how pretraining loss changes as a function of model size, data size,
and training compute. In language modeling, prior work has shown that loss often follows smooth power-law trends with the number of model parameters, training
tokens, and compute \citep{kaplan2020scaling}. The Chinchilla study further showed that compute-efficient training requires scaling model size and data size together, rather than increasing model capacity alone \citep{hoffmann2022training}. We perform an analogous scaling analysis for SensOFormer to understand how masked-reconstruction loss depends on the number of model parameters and the amount of pretraining data.

We fit the validation masked-reconstruction loss using a separable scaling law:
\begin{equation}
    \mathcal{L}(N,D)
    =
    \mathcal{L}_{\infty}
    +
    \frac{A}{N^{\alpha}}
    +
    \frac{B}{D^{\beta}},
    \label{eq:scaling_law}
\end{equation}
where $N$ is the number of trainable model parameters, $D$ is the amount of pretraining data, and $\mathcal{L}(N,D)$ is the validation loss. The constant $\mathcal{L}_{\infty}$ represents the asymptotic loss approached by the fitted model in the limit of large $N$ and $D$. The terms $A/N^{\alpha}$ and $B/D^{\beta}$ represent model-limited and data-limited contributions to the loss, respectively. The exponents $\alpha$ and $\beta$ determine how quickly each contribution decreases as model size or data size increases.

We use this scaling analysis as an empirical diagnostic rather than as an extrapolation guarantee. SensOFormer differs from language models in both data structure and objective: the inputs are heterogeneous accelerator sensor windows rather than text tokens, and the pretraining task is masked denoising reconstruction rather than next-token prediction. Nevertheless, Eq.~\eqref{eq:scaling_law} provides a compact way to summarize whether increasing model capacity and pretraining data improves learned accelerator representations. In our study, we vary $N$ using the five SensOFormer variants in Table~\ref{tab:sensoformer_variants} and vary $D$ by subsampling the pretraining corpus at 12.5\%, 25\%, 50\%, and 100\%. Each scaling run uses single-pass pretraining, and the loss is evaluated on the validation set. The scaling study is performed with an effective batch size of 64 on 32 NVIDIA H100 GPUs.

\begin{figure}[h]
\centering
\includegraphics[width=0.85\textwidth,trim=0cm 0cm 0cm 0cm, clip]{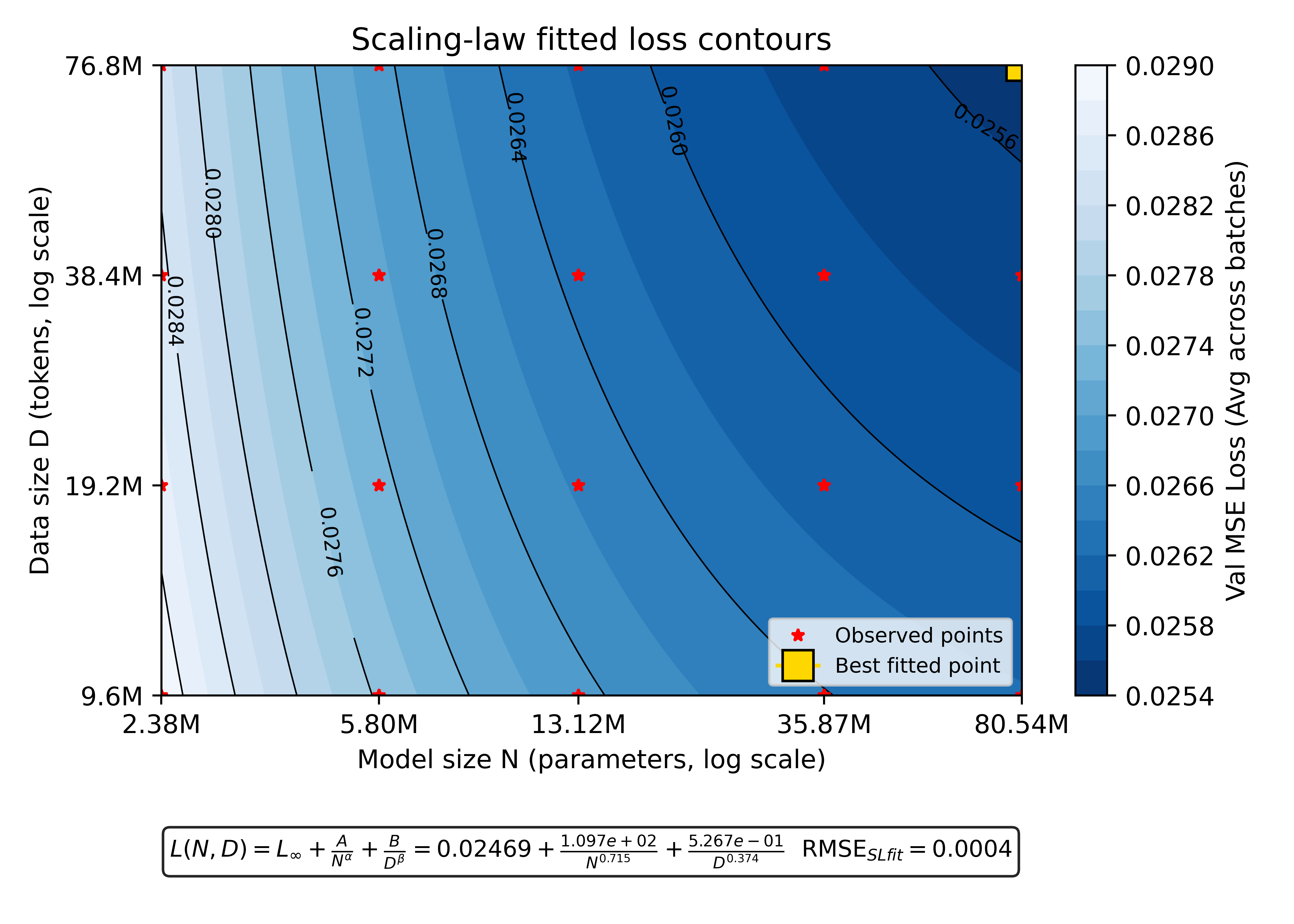}
\caption{Scaling-law fit for SensOFormer validation masked-reconstruction loss as a function of model size $N$ and pretraining data size $D$. Red markers denote
observed model--data configurations, and the yellow marker denotes the best fitted point within the evaluated grid. The fitted surface shows decreasing loss
with both model capacity and data size.}
\label{fig:scaling}
\end{figure}

Fig.~\ref{fig:scaling} shows the fitted scaling surface obtained by nonlinear least squares. The fitted loss is
\begin{equation}
    \mathcal{L}(N,D)
    =
    0.02469
    +
    \frac{1.097\times 10^{2}}{N^{0.715}}
    +
    \frac{5.267\times 10^{-1}}{D^{0.374}},
    \label{eq:fitted_scaling_law}
\end{equation}
with an RMSE of $4\times 10^{-4}$ over the observed model--data configurations. Here, $N$ and $D$ are measured in raw counts rather than in millions. The low RMSE indicates that the separable power law fits the measured validation losses well within the evaluated grid. We do not interpret this as a guarantee that the same law will hold outside the measured range.

The fitted surface shows that validation loss decreases as both model size and data size increase. The best fitted point lies at the largest evaluated model and data scale, indicating that SensOFormer continues to benefit from scale in this regime. The contour geometry also shows that increasing model size produces a strong loss reduction across the evaluated range, especially when moving from the smallest models to the larger S and B variants. Increasing data size also improves loss, but the fitted exponent for data is smaller than that for model size, with $\beta=0.374$ and $\alpha=0.715$. This means that, within the fitted law, the data-limited term decays more slowly with additional data than the model-limited term decays with additional parameters. Thus, although the current evaluated regime still benefits strongly from increasing model capacity, future compute-efficient scaling should increase both model size and pretraining data rather than scaling parameters alone.

We can use the fitted law to derive the compute-optimal allocation of model size and data size \citep{hoffmann2022training}. For a fixed training recipe, let the training compute budget be approximately proportional to the product of model size and data size,
\begin{equation}
    \mathcal{C} = \kappa N D,
\end{equation}
where $\kappa$ absorbs architecture- and implementation-dependent constants. Let $K=\mathcal{C}/\kappa$, so the compute constraint is $ND=K$. Ignoring the constant $\mathcal{L}_{\infty}$, the compute-optimal allocation solves
\begin{equation}
    \min_{N,D}
    \left(
    \frac{A}{N^{\alpha}}
    +
    \frac{B}{D^{\beta}}
    \right)
    \quad
    \mathrm{s.t.}
    \quad
    ND=K.
\end{equation}
Substituting $D=K/N$ and differentiating with respect to $N$ gives the optimality condition
\begin{equation}
    \alpha A (N^\star)^{-\alpha}
    =
    \beta B (D^\star)^{-\beta}.
\end{equation}

This condition states that, at the compute-optimal point, the marginal benefit of increasing model size is balanced against the marginal benefit of increasing
data size under the fixed-compute constraint. Solving for the optimal model and data scales gives
\begin{equation}
    N^\star(K)
    =
    \left(\frac{\alpha A}{\beta B}\right)^{\frac{1}{\alpha+\beta}}
    K^{\frac{\beta}{\alpha+\beta}},
    \qquad
    D^\star(K)
    =
    \left(\frac{\beta B}{\alpha A}\right)^{\frac{1}{\alpha+\beta}}
    K^{\frac{\alpha}{\alpha+\beta}}.
    \label{eq:compute_optimal_scaling}
\end{equation}
Using the fitted exponents from Eq.~\eqref{eq:fitted_scaling_law}, we obtain
\begin{equation}
    N^\star \propto \mathcal{C}^{0.343},
    \qquad
    D^\star \propto \mathcal{C}^{0.657}.
\end{equation}

Under the fitted scaling law, additional compute should therefore be allocated to both larger models and more pretraining data. However, the compute-optimal trajectory increases data size faster than parameter count as the compute budget grows. This is consistent with the broader lesson from Chinchilla-style scaling: larger models are most useful when accompanied by sufficient pretraining data. For SensOFormer, the scaling study suggests that future accelerator foundation models should expand the pretraining corpus alongside model capacity to remain compute-efficient.

\newpage
\section{Extended Results} \label{app:extended_results}


\subsection{LANSCE 2019-2024 data} \label{sec:viz_lansce_1_MDI}
\subsubsection{Point-wise missing data imputation}

\begin{figure}[h]
\centering
\begin{minipage}{1.0\textwidth}
  \includegraphics[width=0.95\linewidth,trim=0cm 0cm 0cm 0cm,clip]{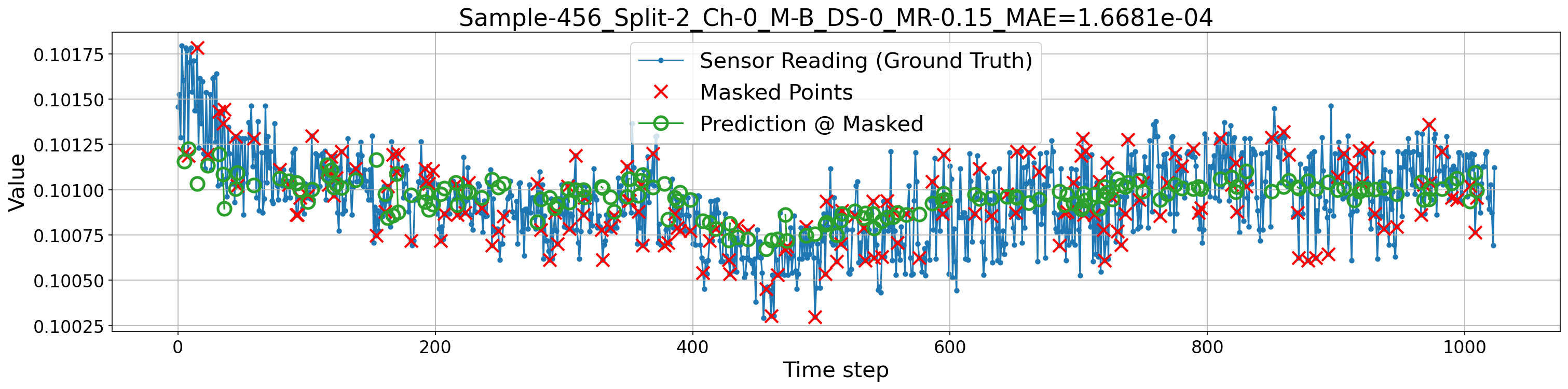}
\end{minipage}
\begin{minipage}{1.0\textwidth}
  \includegraphics[width=0.95\linewidth,trim=0cm 0cm 0cm 0cm,clip]{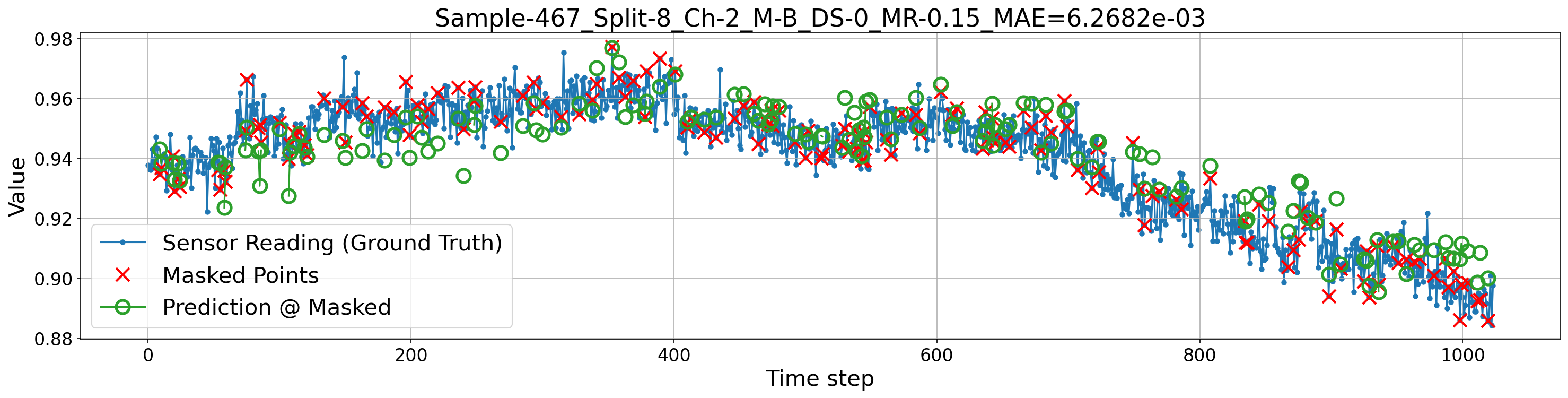}
\end{minipage}
\begin{minipage}{1.0\textwidth}
  \includegraphics[width=0.95\linewidth,trim=0cm 0cm 0cm 0cm,clip]{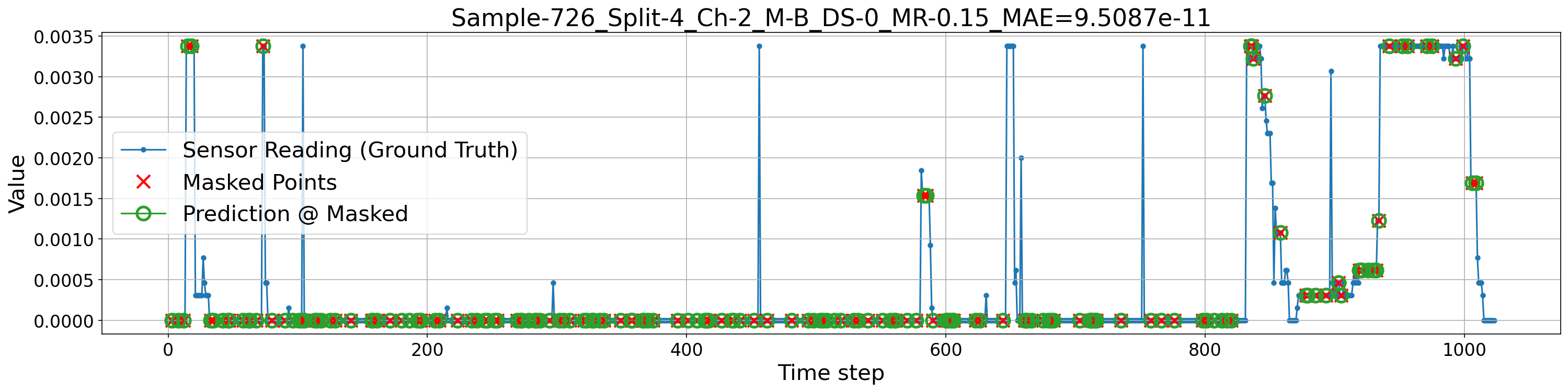}
\end{minipage}
\begin{minipage}{1.0\textwidth}
  \includegraphics[width=0.95\linewidth,trim=0cm 0cm 0cm 0cm,clip]{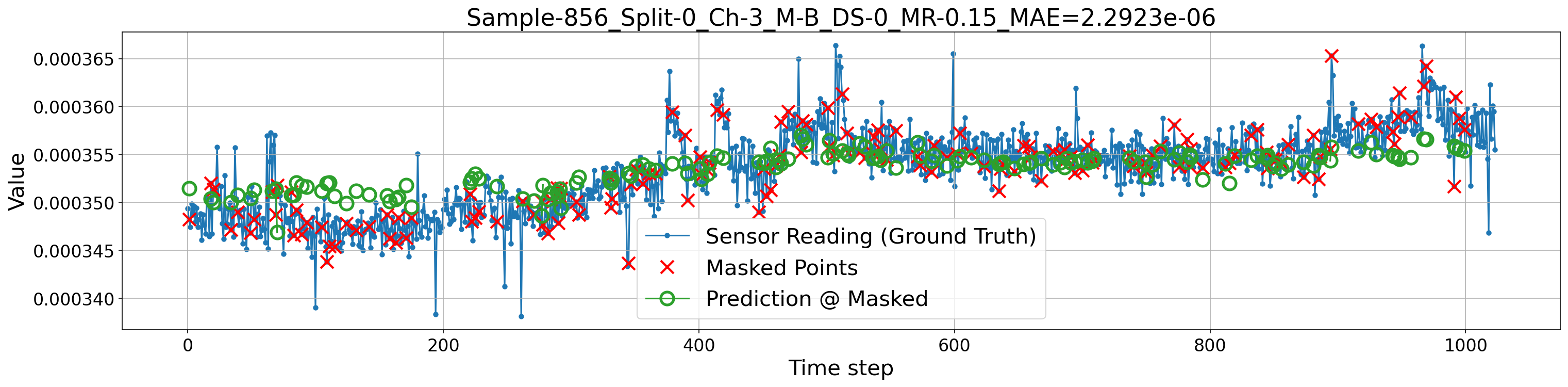}
\end{minipage}
\begin{minipage}{1.0\textwidth}
  \includegraphics[width=0.95\linewidth,trim=0cm 0cm 0cm 0cm,clip]{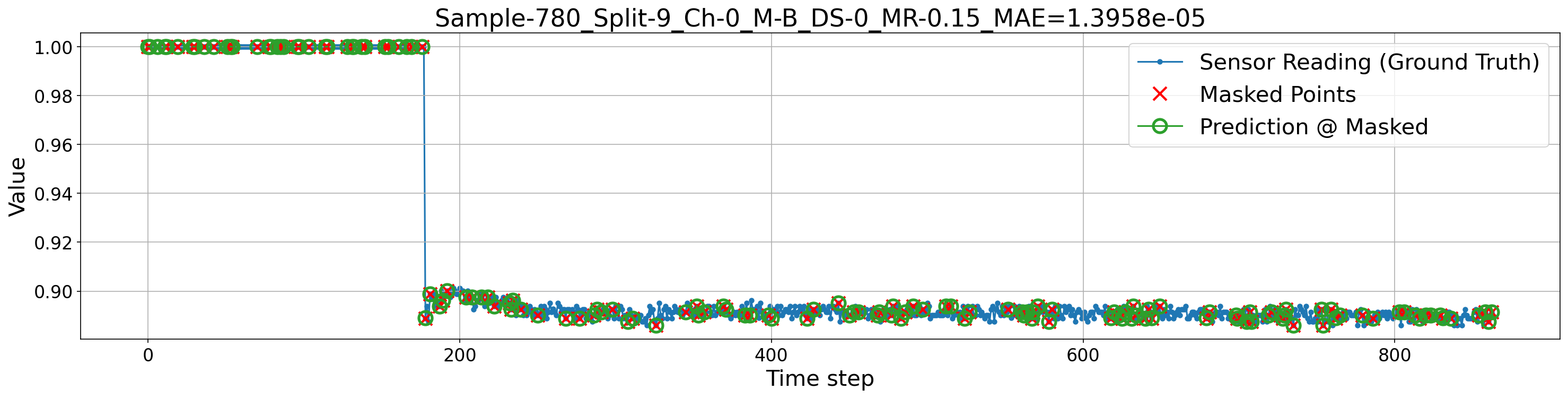}
\end{minipage}
\caption{Representative zero-shot \emph{point-wise imputation} examples from pretrained \emph{SensOFormer-B} on held-out \emph{LANSCE 2019--2024} test windows with a \emph{15\% mask ratio}. Blue curves show the normalized ground-truth sensor values, red markers indicate masked entries, and green circles show the model predictions at the masked locations. Each panel corresponds to a different test window and channel. The subplot title reports the sample index, split, channel, model size (M), dataset (DS), mask ratio (MR), and mean absolute error (MAE) computed on the masked entries.}
\label{fig:viz_mdi_point_ds0}
\end{figure}

\newpage
\subsubsection{Block-wise missing data imputation}
\begin{figure}[h]
\centering
\begin{minipage}{1.0\textwidth}
  \includegraphics[width=0.95\linewidth,trim=0cm 0cm 0cm 0cm,clip]{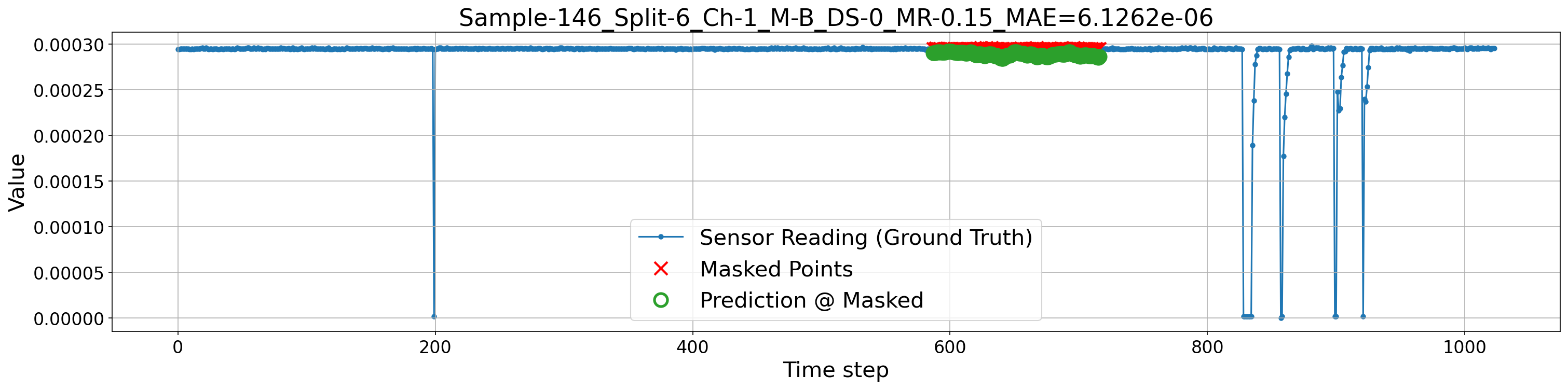}
\end{minipage}
\begin{minipage}{1.0\textwidth}
  \includegraphics[width=0.95\linewidth,trim=0cm 0cm 0cm 0cm,clip]{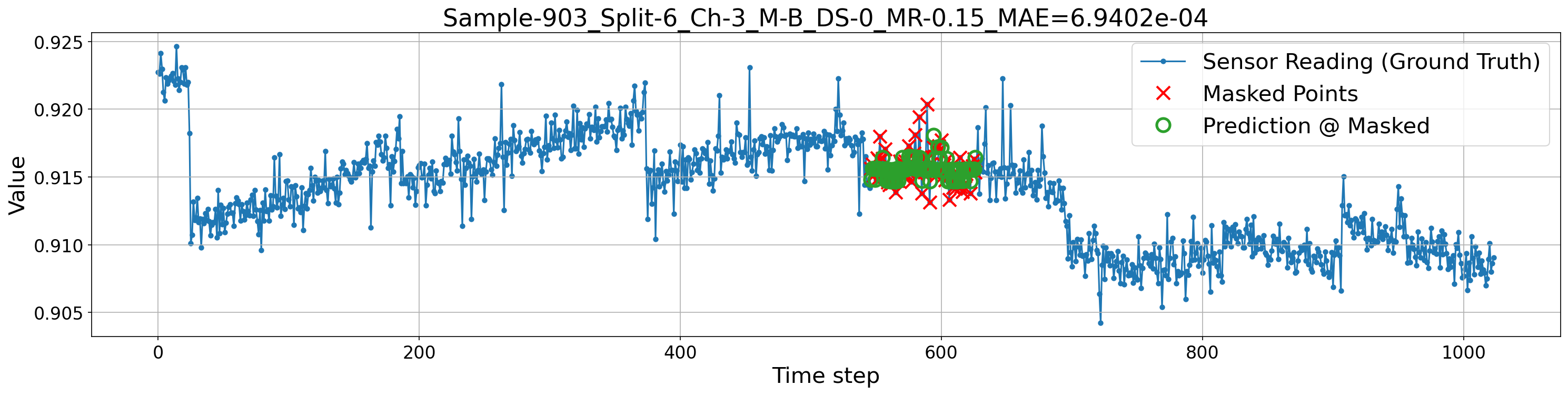}
\end{minipage}
\begin{minipage}{1.0\textwidth}
  \includegraphics[width=0.95\linewidth,trim=0cm 0cm 0cm 0cm,clip]{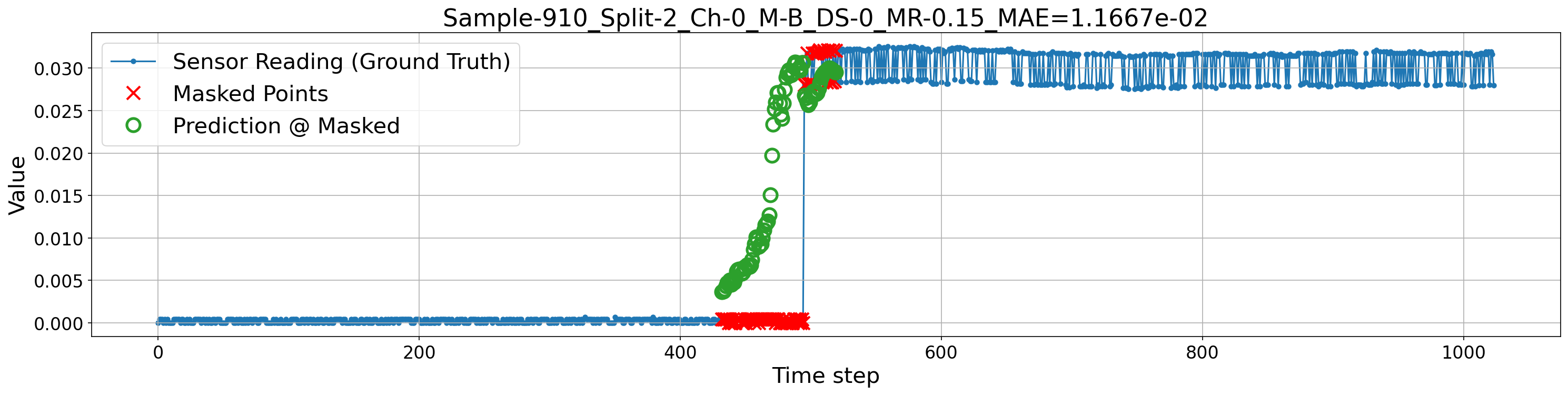}
\end{minipage}
\begin{minipage}{1.0\textwidth}
  \includegraphics[width=0.95\linewidth,trim=0cm 0cm 0cm 0cm,clip]{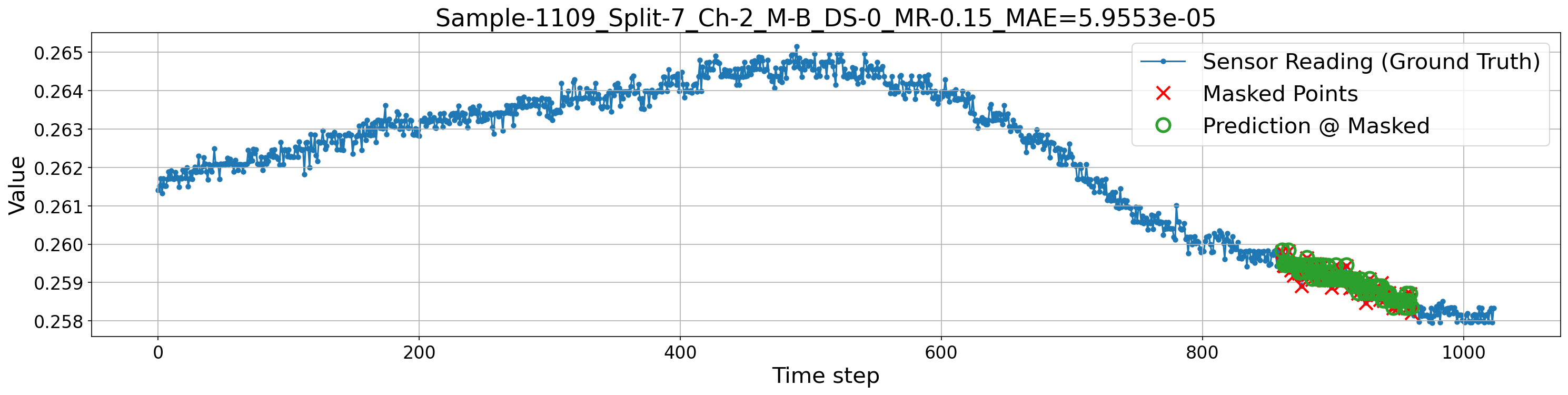}
\end{minipage}
\begin{minipage}{1.0\textwidth}
  \includegraphics[width=0.95\linewidth,trim=0cm 0cm 0cm 0cm,clip]{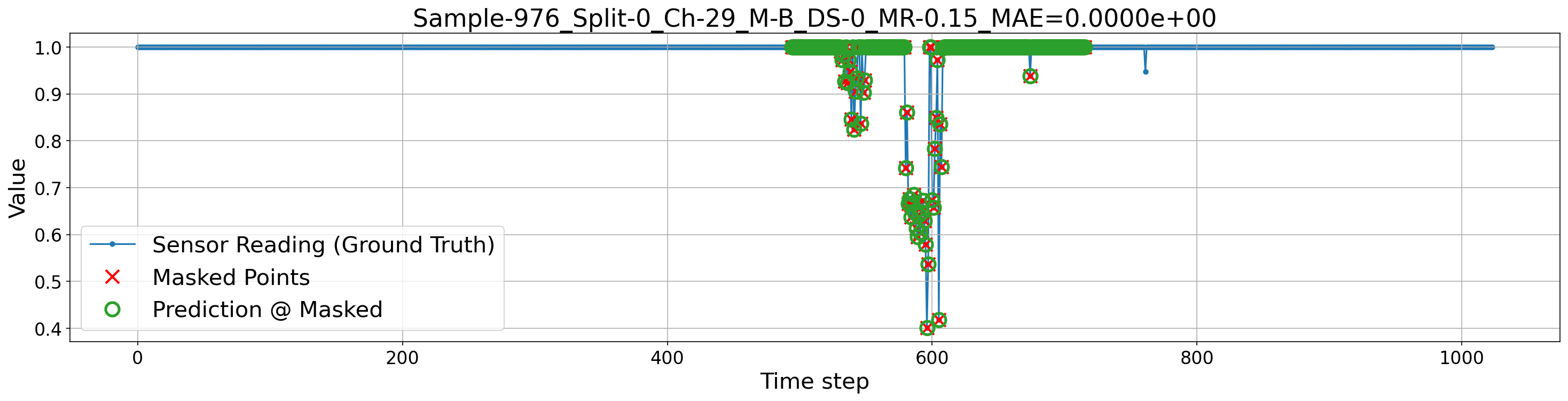}
\end{minipage}
\caption{Representative zero-shot \emph{block-wise imputation} examples from pretrained \emph{SensOFormer-B} on held-out \emph{LANSCE 2019--2024} test windows with a \emph{15\% mask ratio}. Blue curves show the normalized ground-truth sensor values, red markers indicate masked entries, and green circles show the model predictions at the masked locations. Each panel corresponds to a different test window and channel. The subplot title reports the sample index, split, channel, model size (M), dataset (DS), mask ratio (MR), and mean absolute error (MAE) computed on the masked entries.}
\label{fig:viz_mdi_block_ds0}
\end{figure}


\newpage
\subsection{LANSCE 2025 data} \label{sec:viz_lansce_2_MDI}
\subsubsection{Point-wise missing data imputation}
\begin{figure}[h]
\centering
\begin{minipage}{1.0\textwidth}
  \includegraphics[width=0.95\linewidth,trim=0cm 0cm 0cm 0cm,clip]{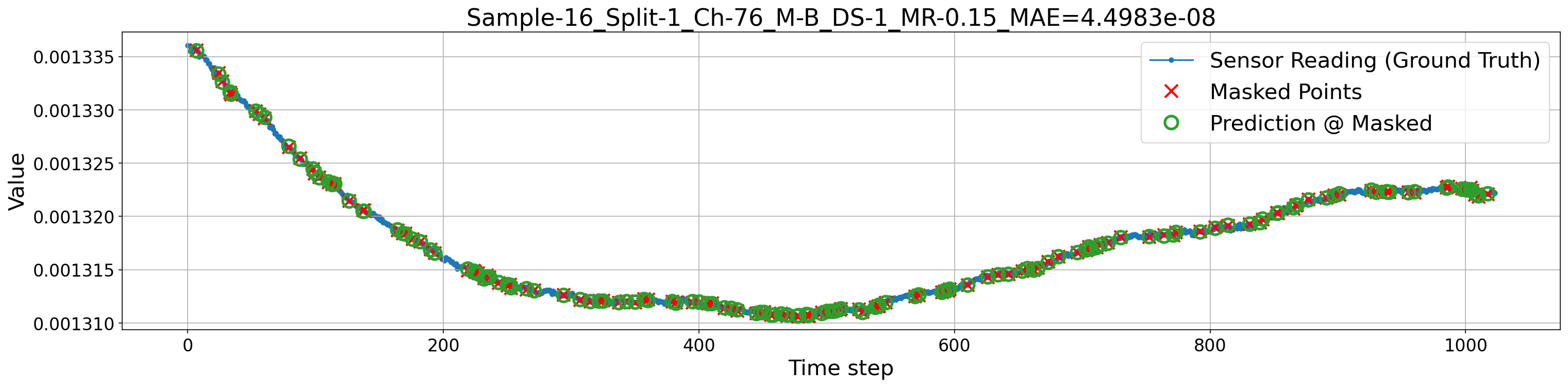}
\end{minipage}
\begin{minipage}{1.0\textwidth}
  \includegraphics[width=0.95\linewidth,trim=0cm 0cm 0cm 0cm,clip]{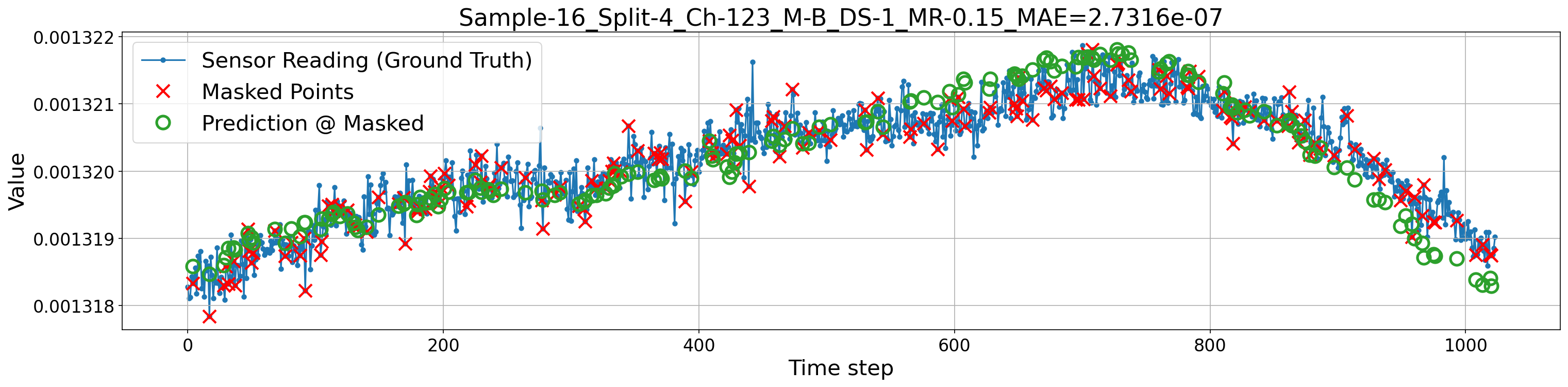}
\end{minipage}
\begin{minipage}{1.0\textwidth}
  \includegraphics[width=0.95\linewidth,trim=0cm 0cm 0cm 0cm,clip]{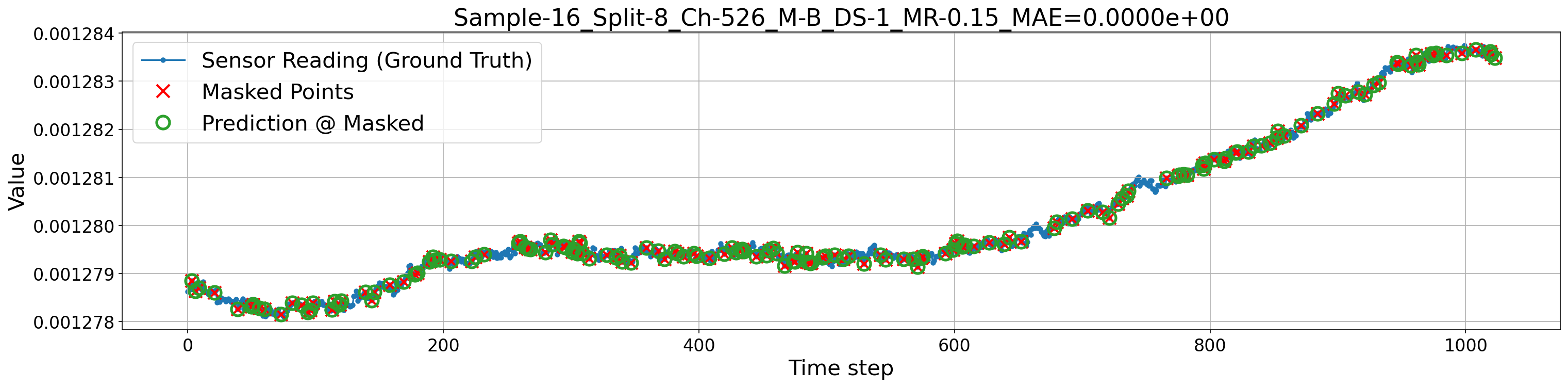}
\end{minipage}
\begin{minipage}{1.0\textwidth}
  \includegraphics[width=0.95\linewidth,trim=0cm 0cm 0cm 0cm,clip]{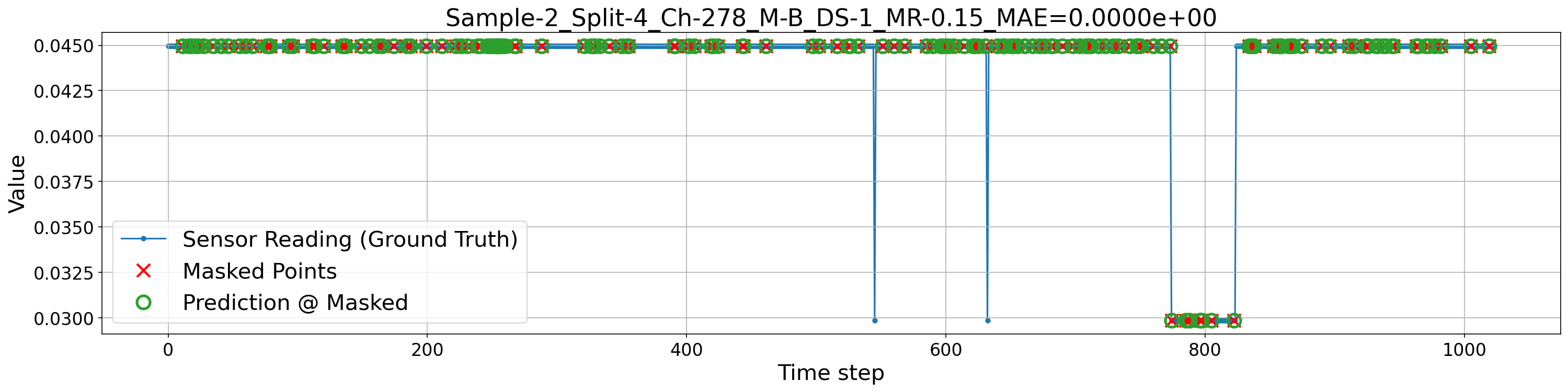}
\end{minipage}
\begin{minipage}{1.0\textwidth}
  \includegraphics[width=0.95\linewidth,trim=0cm 0cm 0cm 0cm,clip]{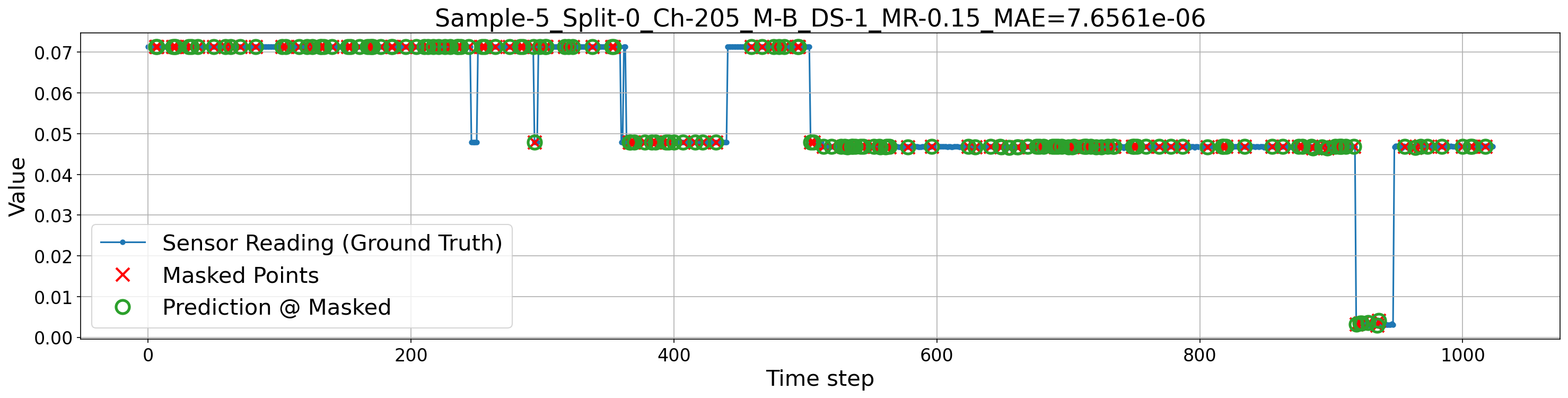}
\end{minipage}
\caption{Representative zero-shot \emph{point-wise imputation} examples from pretrained \emph{SensOFormer-B} on held-out \emph{LANSCE 2025} test windows with a \emph{15\% mask ratio}. Blue curves show the normalized ground-truth sensor values, red markers indicate masked entries, and green circles show the model predictions at the masked locations. Each panel corresponds to a different test window and channel. The subplot title reports the sample index, split, channel, model size (M), dataset (DS), mask ratio (MR), and mean absolute error (MAE) computed on the masked entries.}
\label{fig:viz_mdi_point_ds1}
\end{figure}

\newpage
\subsubsection{Block-wise missing data imputation}
\begin{figure}[h]
\centering
\begin{minipage}{1.0\textwidth}
  \includegraphics[width=0.95\linewidth,trim=0cm 0cm 0cm 0cm,clip]{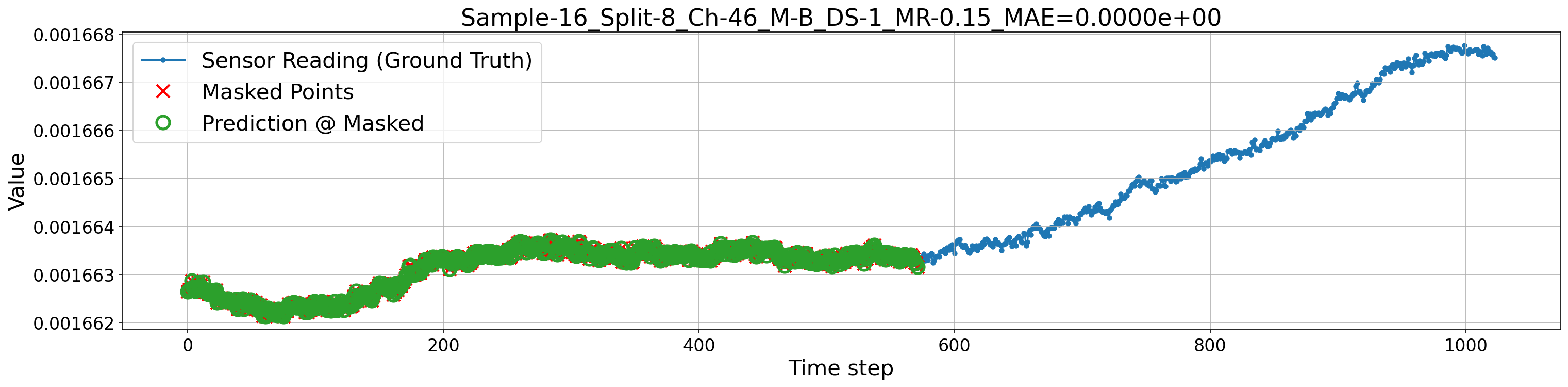}
\end{minipage}
\begin{minipage}{1.0\textwidth}
  \includegraphics[width=0.95\linewidth,trim=0cm 0cm 0cm 0cm,clip]{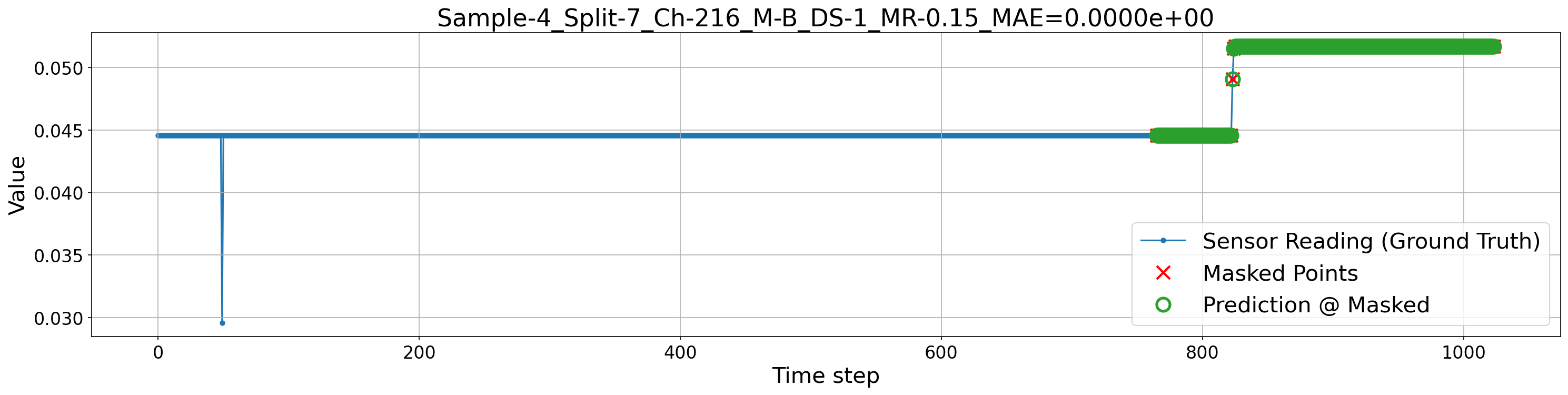}
\end{minipage}
\begin{minipage}{1.0\textwidth}
  \includegraphics[width=0.95\linewidth,trim=0cm 0cm 0cm 0cm,clip]{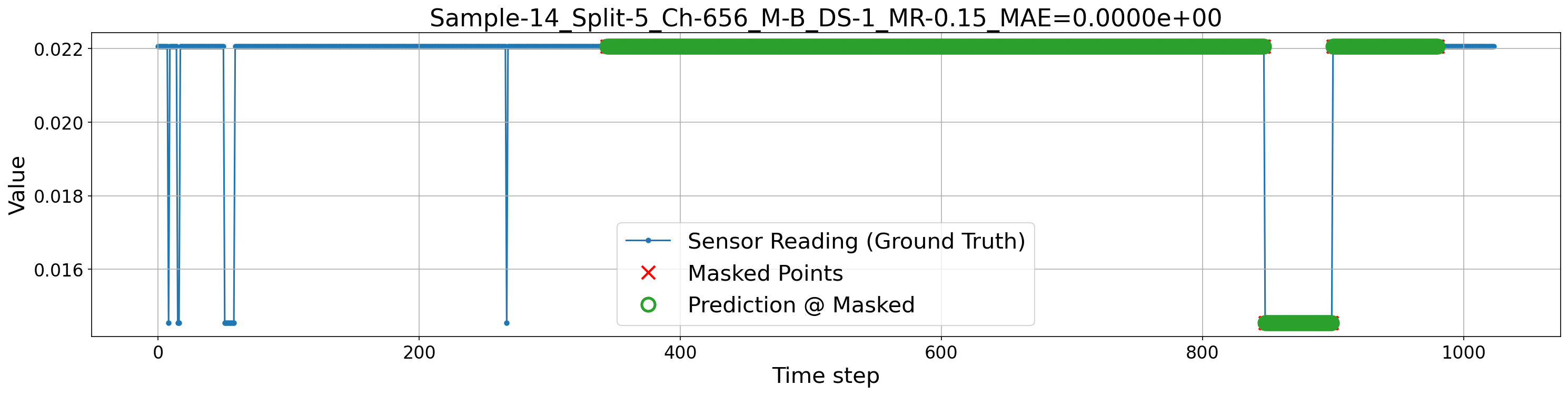}
\end{minipage}
\begin{minipage}{1.0\textwidth}
  \includegraphics[width=0.95\linewidth,trim=0cm 0cm 0cm 0cm,clip]{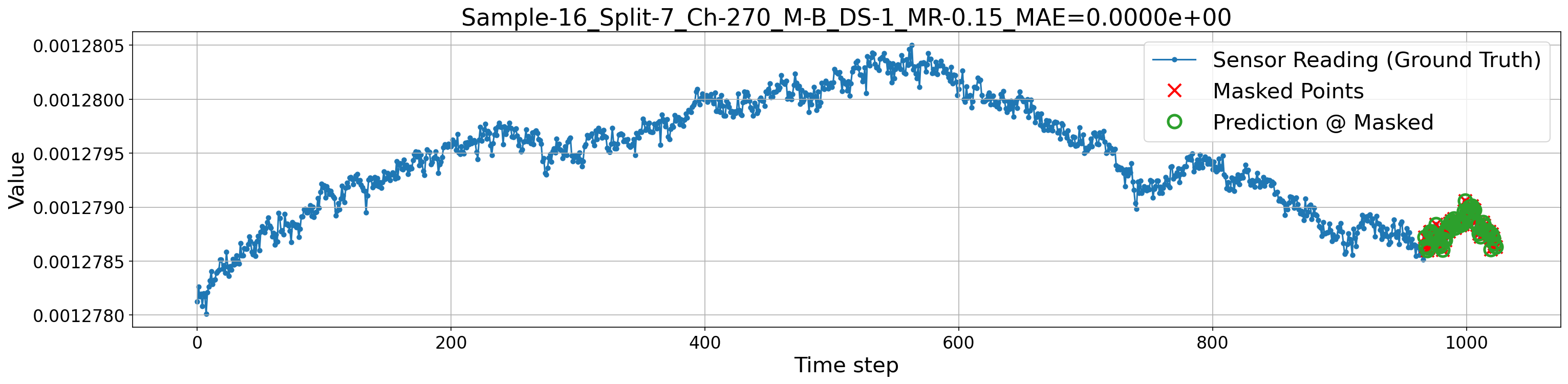}
\end{minipage}
\begin{minipage}{1.0\textwidth}
  \includegraphics[width=0.95\linewidth,trim=0cm 0cm 0cm 0cm,clip]{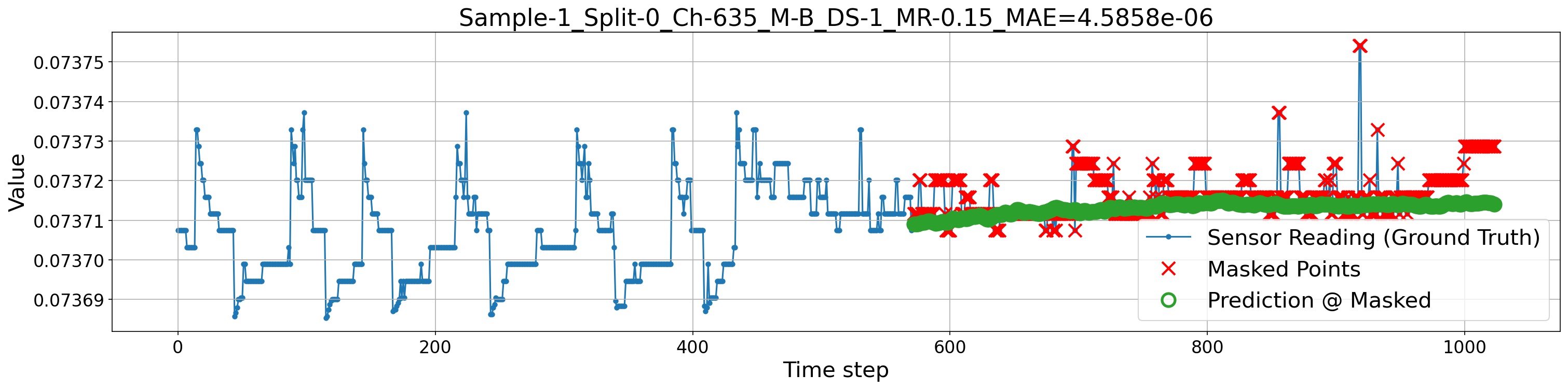}
\end{minipage}
\caption{Representative zero-shot \emph{block-wise imputation} examples from pretrained \emph{SensOFormer-B} on held-out \emph{LANSCE 2025} test windows with a \emph{15\% mask ratio}. Blue curves show the normalized ground-truth sensor values, red markers indicate masked entries, and green circles show the model predictions at the masked locations. Each panel corresponds to a different test window and channel. The subplot title reports the sample index, split, channel, model size (M), dataset (DS), mask ratio (MR), and mean absolute error (MAE) computed on the masked entries.}
\label{fig:viz_mdi_block_ds1}
\end{figure}

\newpage
\subsection{LHC-BL data} \label{sec:viz_lhc_MDI}

\subsubsection{Point-wise missing data imputation}
\begin{figure}[h]
\centering
\begin{minipage}{1.0\textwidth}
  \includegraphics[width=0.95\linewidth,trim=0cm 0cm 0cm 0cm,clip]{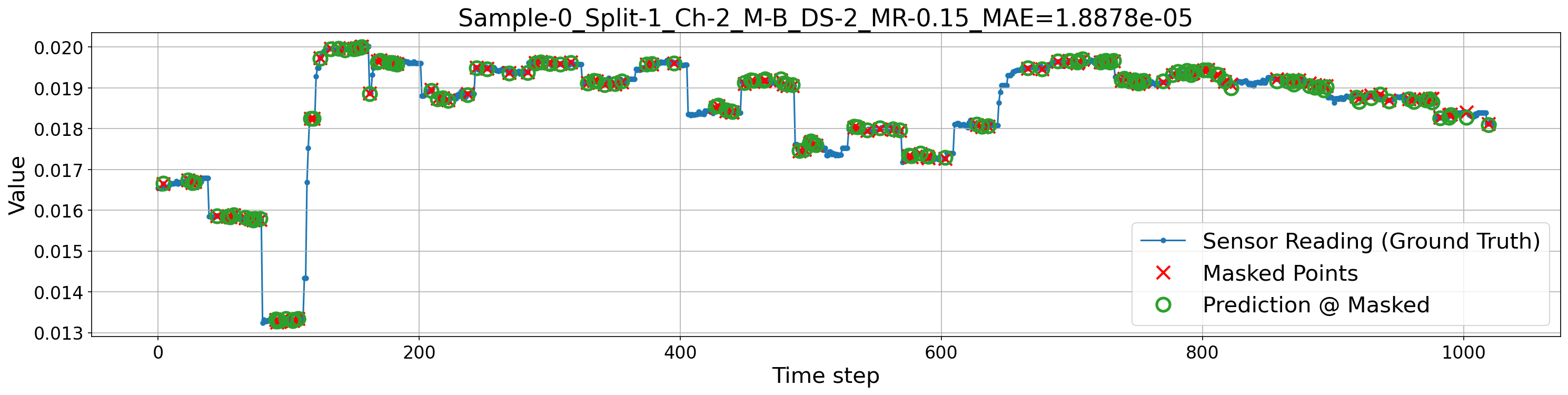}
\end{minipage}
\begin{minipage}{1.0\textwidth}
  \includegraphics[width=0.95\linewidth,trim=0cm 0cm 0cm 0cm,clip]{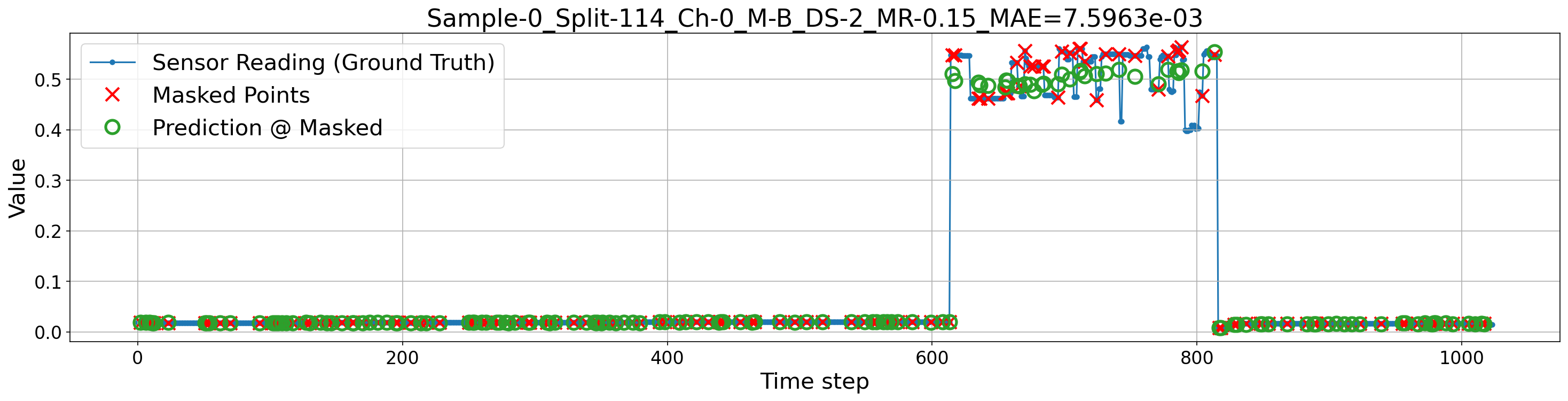}
\end{minipage}
\begin{minipage}{1.0\textwidth}
  \includegraphics[width=0.95\linewidth,trim=0cm 0cm 0cm 0cm,clip]{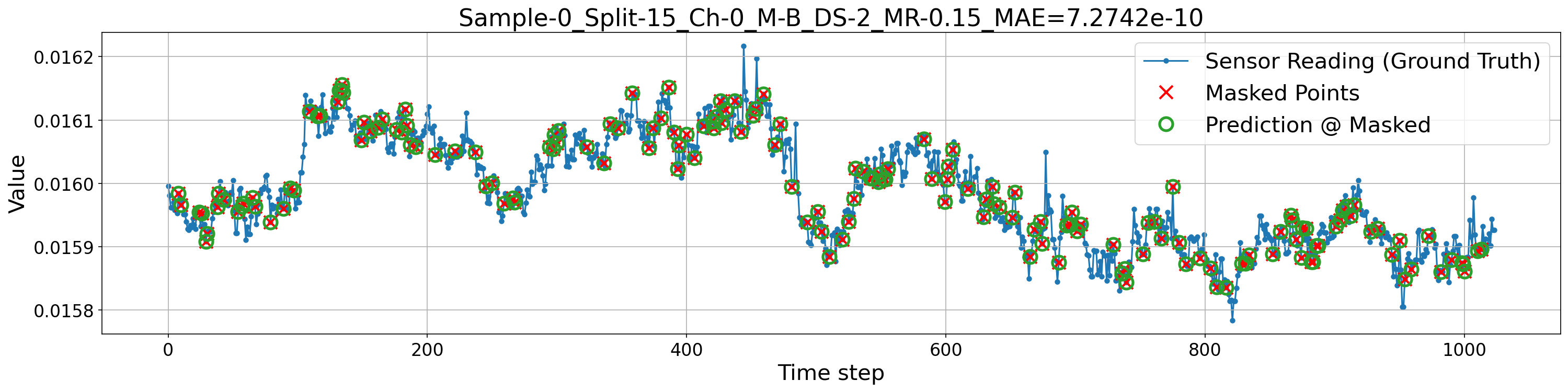}
\end{minipage}
\begin{minipage}{1.0\textwidth}
  \includegraphics[width=0.95\linewidth,trim=0cm 0cm 0cm 0cm,clip]{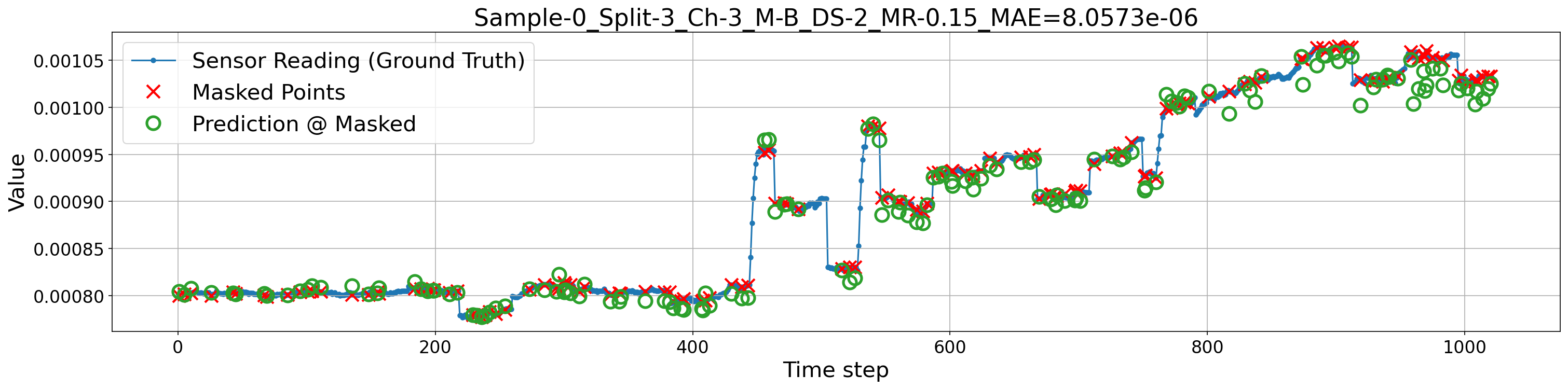}
\end{minipage}
\begin{minipage}{1.0\textwidth}
  \includegraphics[width=0.95\linewidth,trim=0cm 0cm 0cm 0cm,clip]{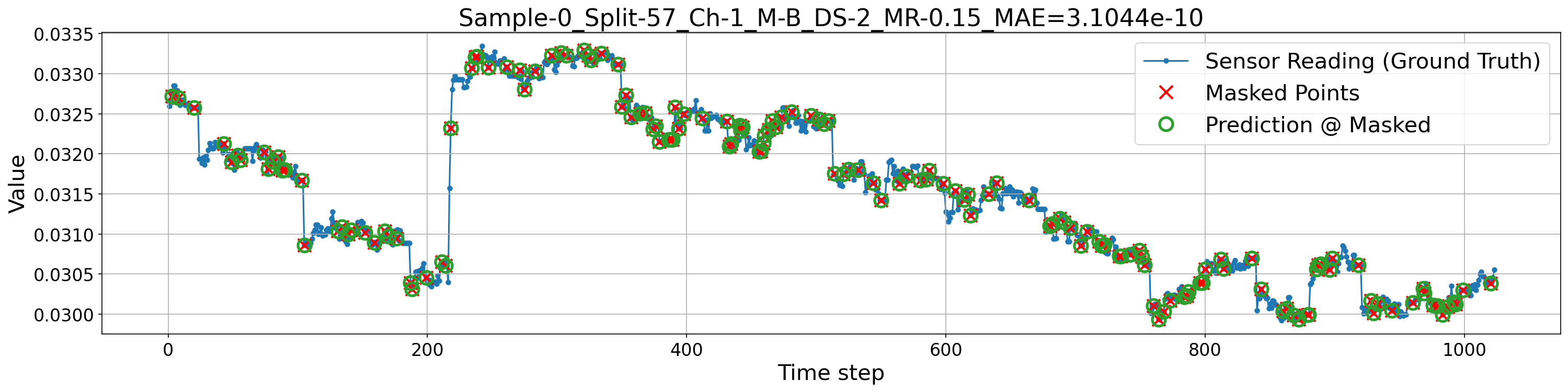}
\end{minipage}
\caption{Representative zero-shot \emph{point-wise imputation} examples from pretrained \emph{SensOFormer-B} on held-out \emph{LHC-BL} test windows with a \emph{15\% mask ratio}. Blue curves show the normalized ground-truth sensor values, red markers indicate masked entries, and green circles show the model predictions at the masked locations. Each panel corresponds to a different test window and channel. The subplot title reports the sample index, split, channel, model size (M), dataset (DS), mask ratio (MR), and mean absolute error (MAE) computed on the masked entries.}
\label{fig:viz_mdi_point_ds2}
\end{figure}


\newpage
\subsection{HVCM-SNS data} \label{sec:viz_sns_MDI}

\subsubsection{Point-wise missing data imputation}
\begin{figure}[h]
\centering
\begin{minipage}{1.0\textwidth}
  \includegraphics[width=0.95\linewidth,trim=0cm 0cm 0cm 0cm,clip]{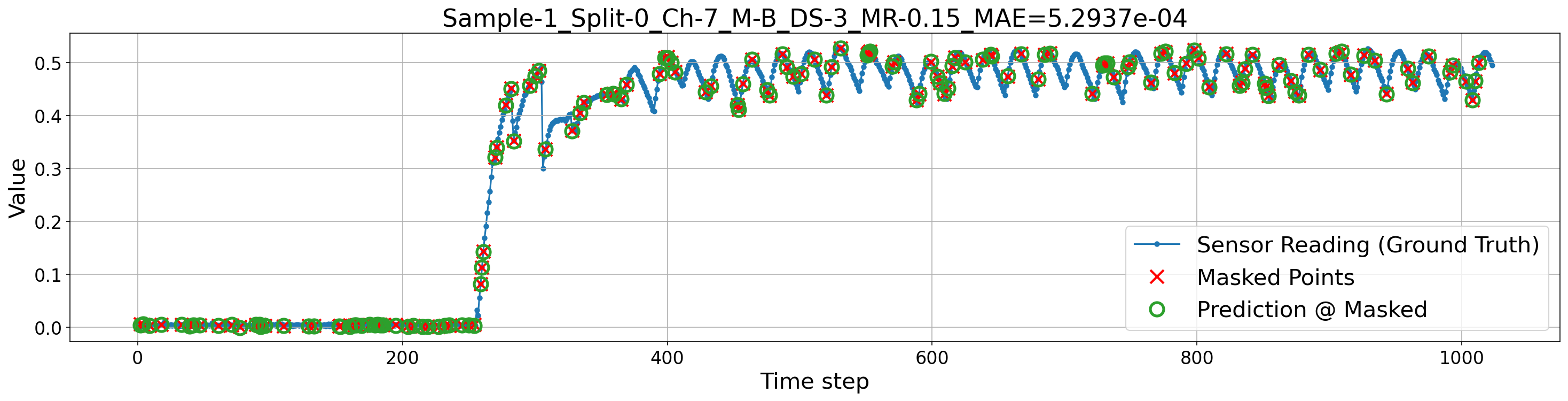}
\end{minipage}
\begin{minipage}{1.0\textwidth}
  \includegraphics[width=0.95\linewidth,trim=0cm 0cm 0cm 0cm,clip]{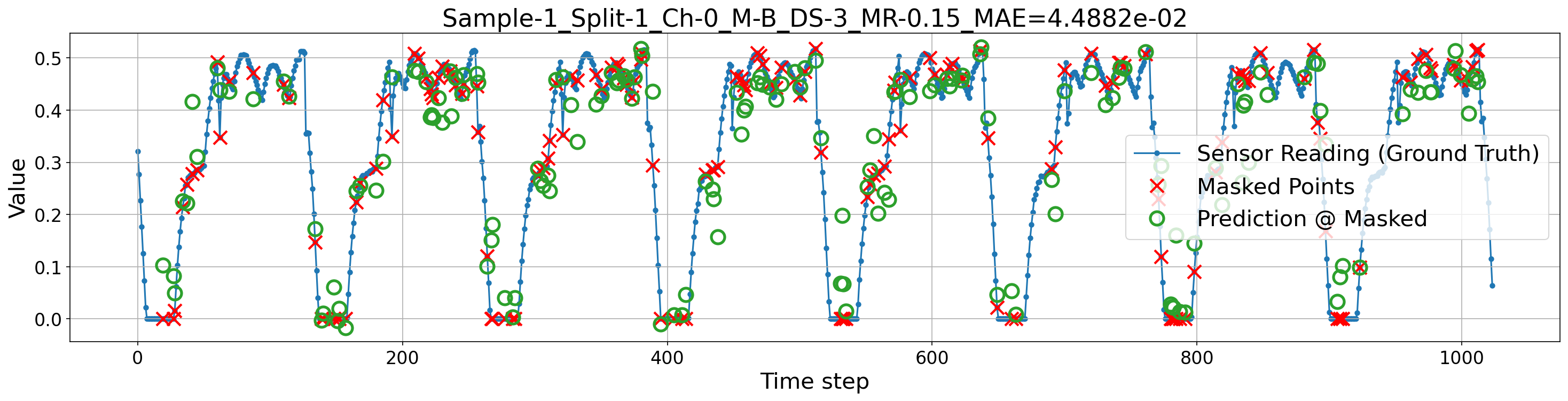}
\end{minipage}
\begin{minipage}{1.0\textwidth}
  \includegraphics[width=0.95\linewidth,trim=0cm 0cm 0cm 0cm,clip]{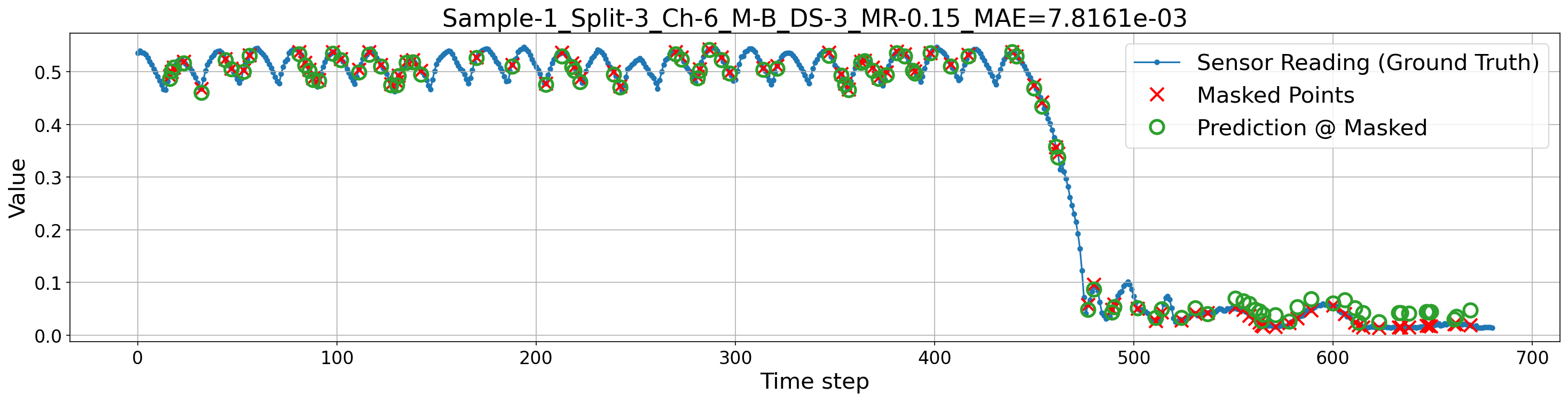}
\end{minipage}
\begin{minipage}{1.0\textwidth}
  \includegraphics[width=0.95\linewidth,trim=0cm 0cm 0cm 0cm,clip]{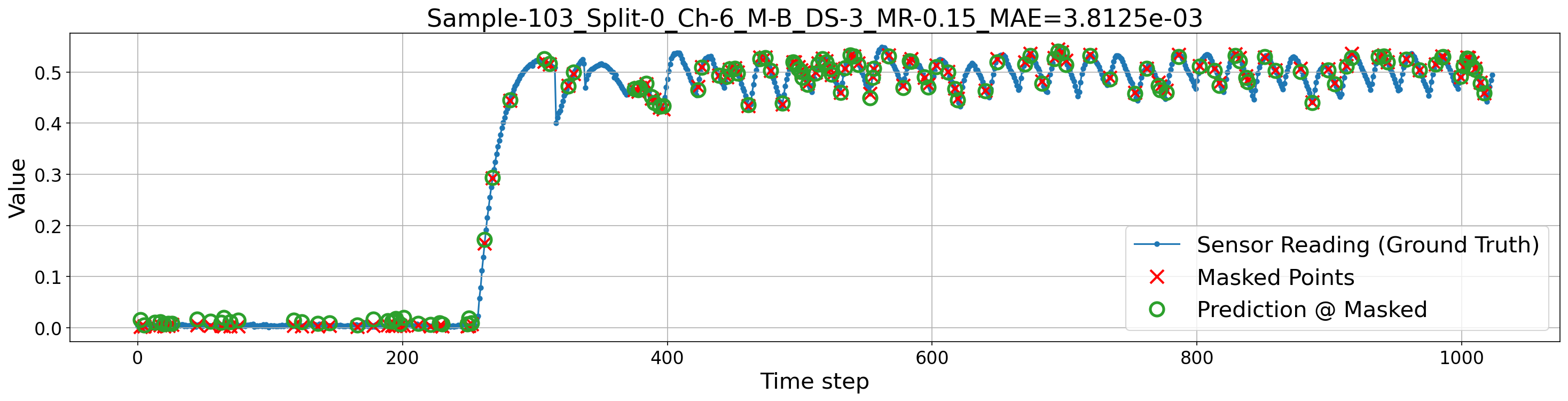}
\end{minipage}
\begin{minipage}{1.0\textwidth}
  \includegraphics[width=0.95\linewidth,trim=0cm 0cm 0cm 0cm,clip]{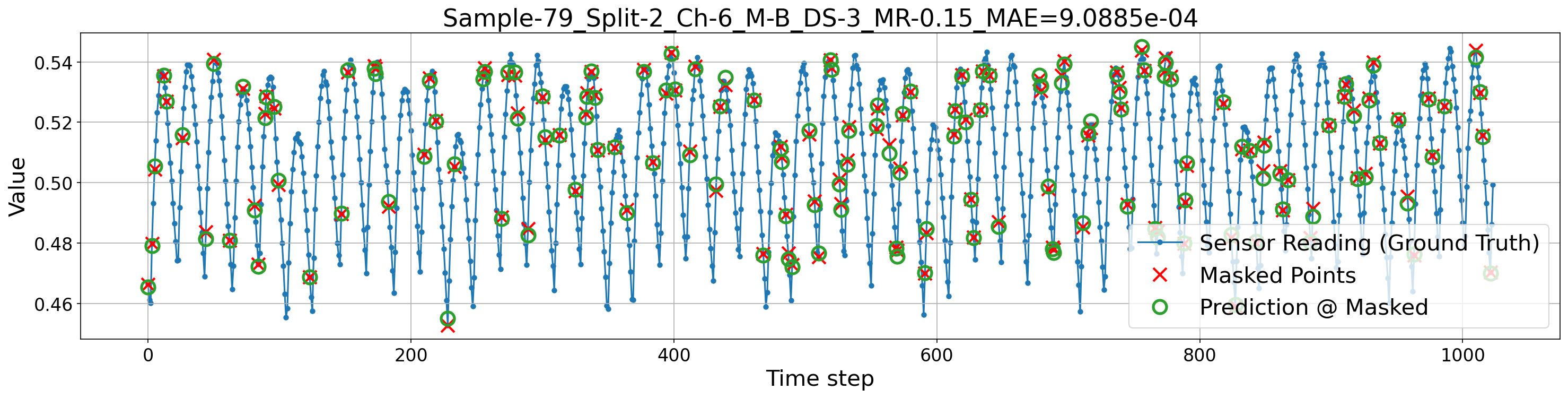}
\end{minipage}
\caption{Representative zero-shot \emph{point-wise imputation} examples from pretrained \emph{SensOFormer-B} on held-out \emph{HVCM-SNS} test windows with a \emph{15\% mask ratio}. Blue curves show the normalized ground-truth sensor values, red markers indicate masked entries, and green circles show the model predictions at the masked locations. Each panel corresponds to a different test window and channel. The subplot title reports the sample index, split, channel, model size (M), dataset (DS), mask ratio (MR), and mean absolute error (MAE) computed on the masked entries.}
\label{fig:viz_mdi_point_ds3}
\end{figure}

\newpage
\subsubsection{Block-wise missing data imputation}
\begin{figure}[h]
\centering
\begin{minipage}{1.0\textwidth}
  \includegraphics[width=0.95\linewidth,trim=0cm 0cm 0cm 0cm,clip]{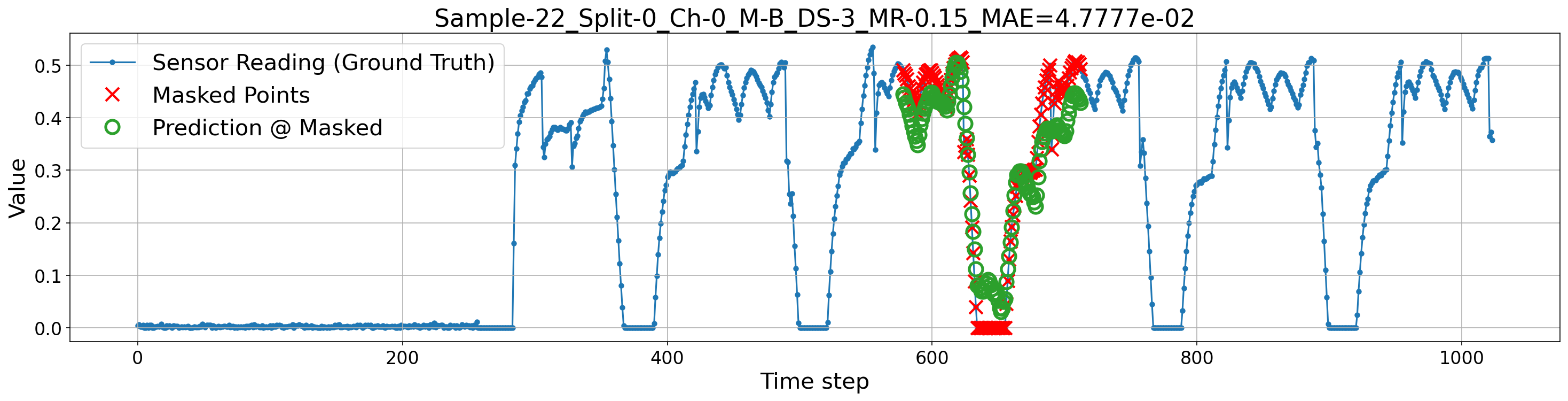}
\end{minipage}
\begin{minipage}{1.0\textwidth}
  \includegraphics[width=0.95\linewidth,trim=0cm 0cm 0cm 0cm,clip]{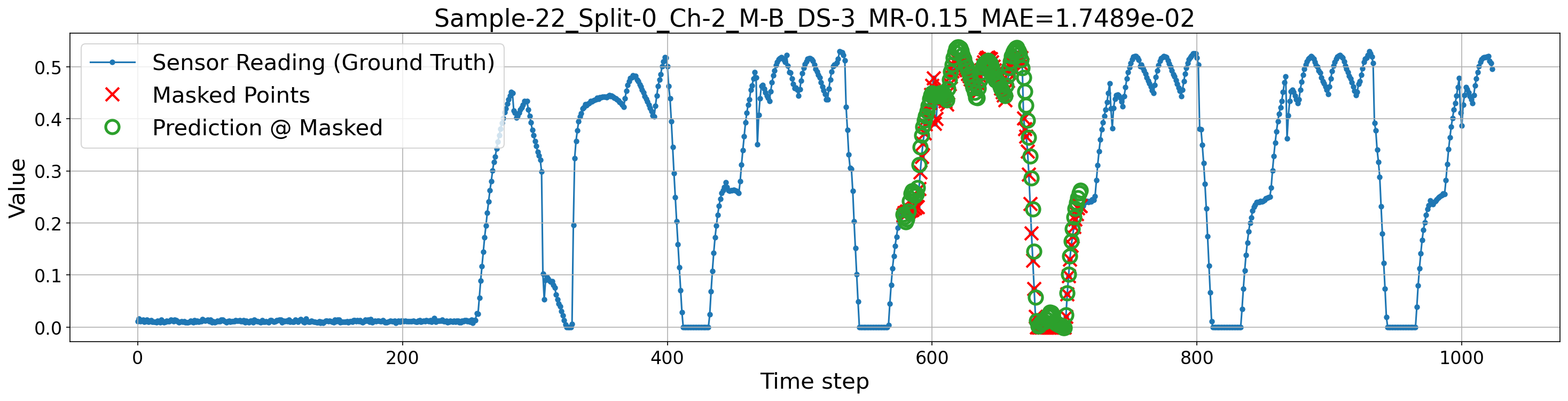}
\end{minipage}
\begin{minipage}{1.0\textwidth}
  \includegraphics[width=0.95\linewidth,trim=0cm 0cm 0cm 0cm,clip]{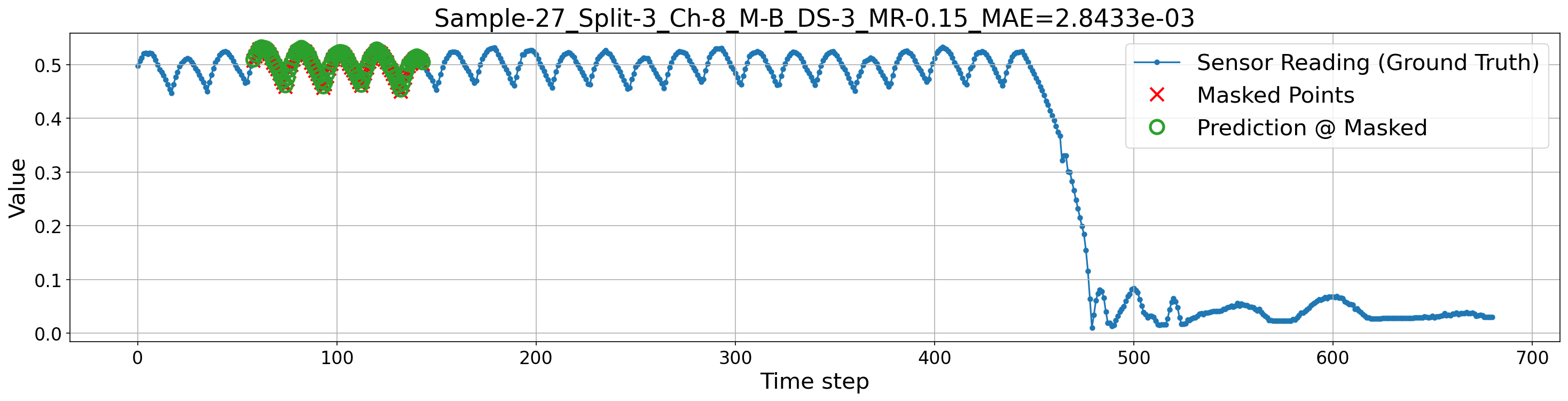}
\end{minipage}
\begin{minipage}{1.0\textwidth}
  \includegraphics[width=0.95\linewidth,trim=0cm 0cm 0cm 0cm,clip]{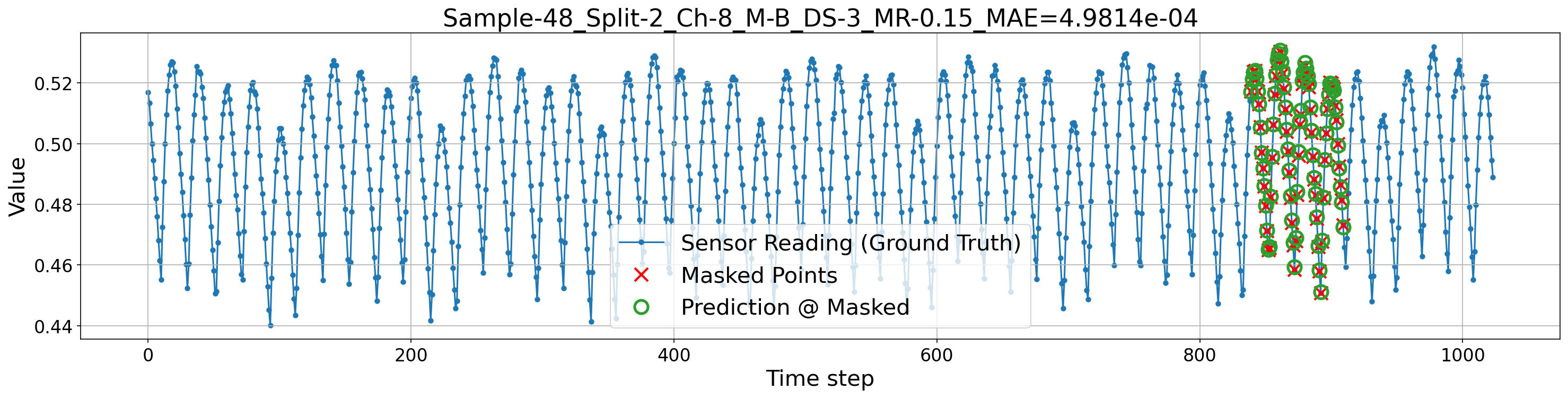}
\end{minipage}
\begin{minipage}{1.0\textwidth}
  \includegraphics[width=0.95\linewidth,trim=0cm 0cm 0cm 0cm,clip]{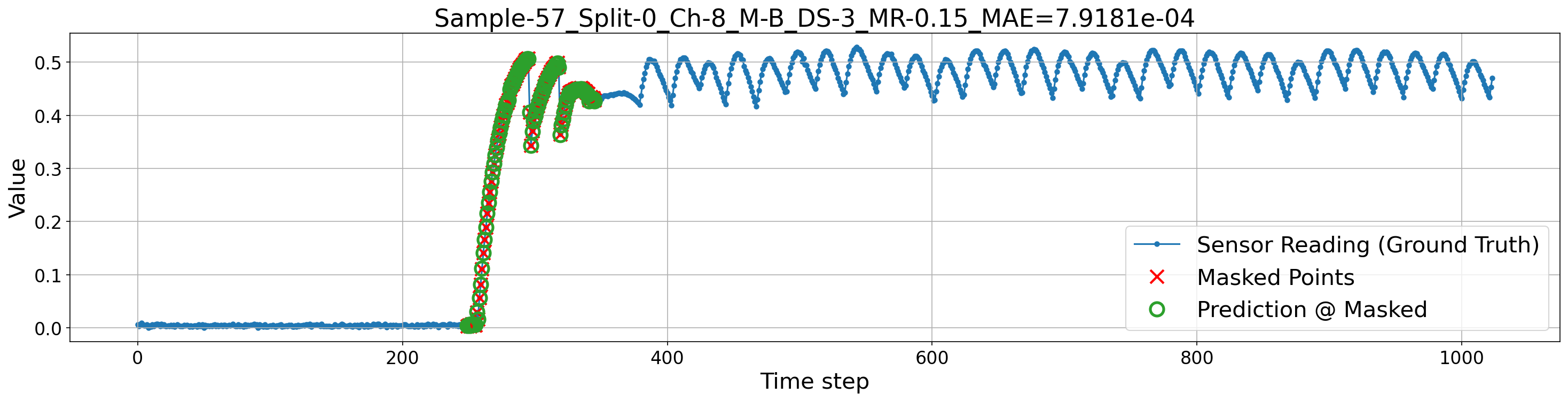}
\end{minipage}
\caption{Representative zero-shot \emph{block-wise imputation} examples from pretrained \emph{SensOFormer-B} on held-out \emph{HVCM-SNS} test windows with a \emph{15\% mask ratio}. Blue curves show the normalized ground-truth sensor values, red markers indicate masked entries, and green circles show the model predictions at the masked locations. Each panel corresponds to a different test window and channel. The subplot title reports the sample index, split, channel, model size (M), dataset (DS), mask ratio (MR), and mean absolute error (MAE) computed on the masked entries.}
\label{fig:viz_mdi_block_ds3}
\end{figure}

\newpage
\subsection{RFS-LCLS data} \label{sec:viz_lcls_MDI}

\subsubsection{Point-wise missing data imputation}
\begin{figure}[h]
\centering
\begin{minipage}{1.0\textwidth}
  \includegraphics[width=0.95\linewidth,trim=0cm 0cm 0cm 0cm,clip]{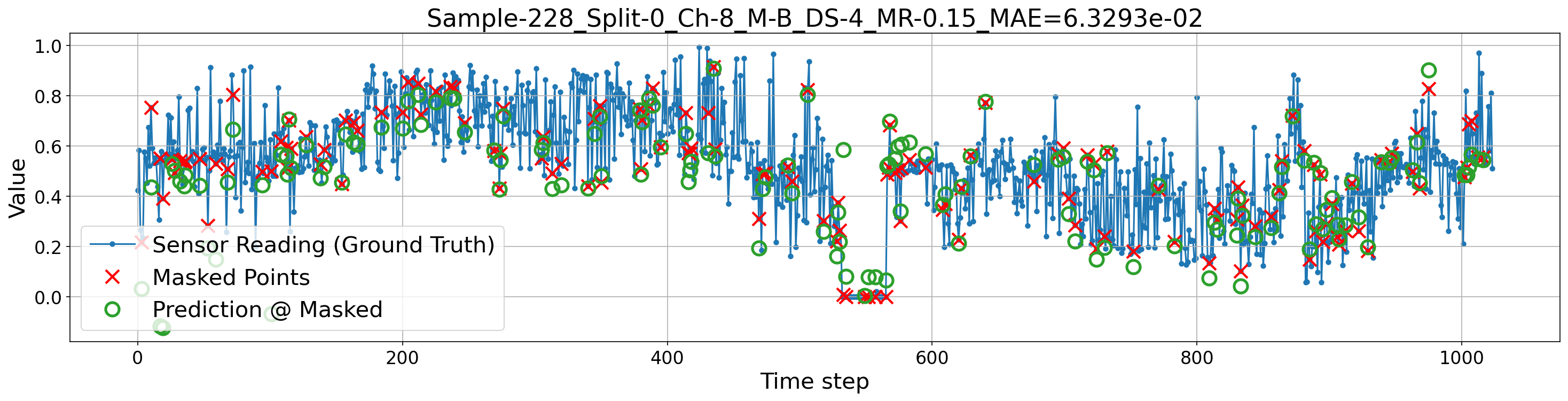}
\end{minipage}
\begin{minipage}{1.0\textwidth}
  \includegraphics[width=0.95\linewidth,trim=0cm 0cm 0cm 0cm,clip]{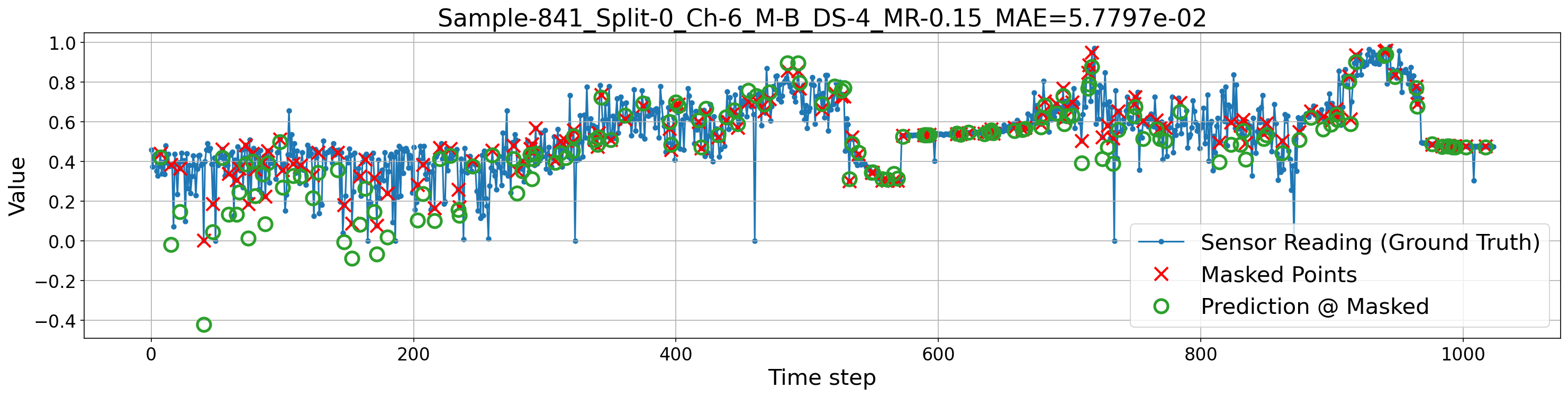}
\end{minipage}
\begin{minipage}{1.0\textwidth}
  \includegraphics[width=0.95\linewidth,trim=0cm 0cm 0cm 0cm,clip]{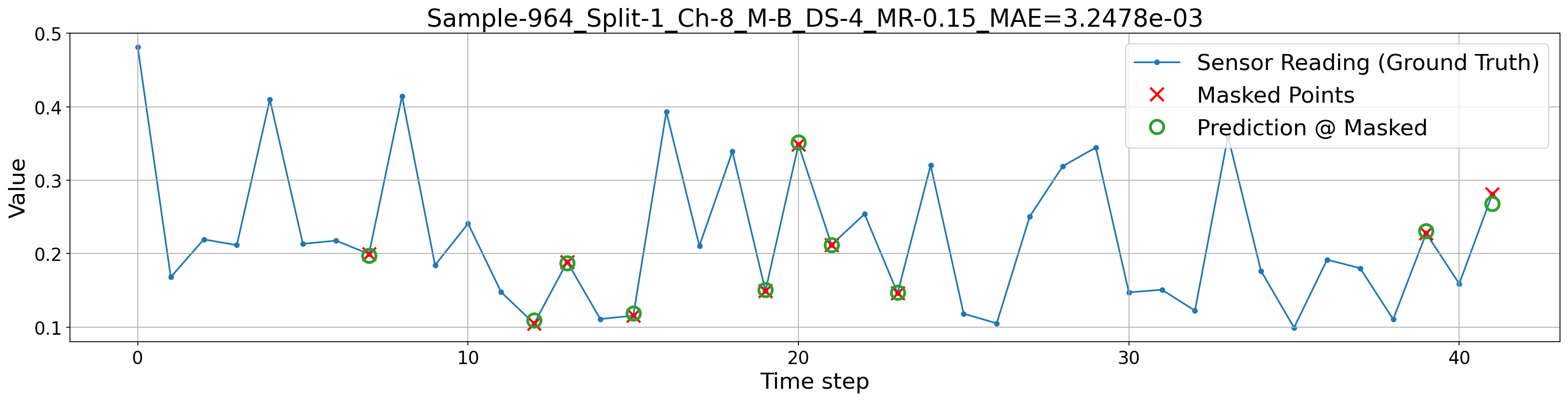}
\end{minipage}
\begin{minipage}{1.0\textwidth}
  \includegraphics[width=0.95\linewidth,trim=0cm 0cm 0cm 0cm,clip]{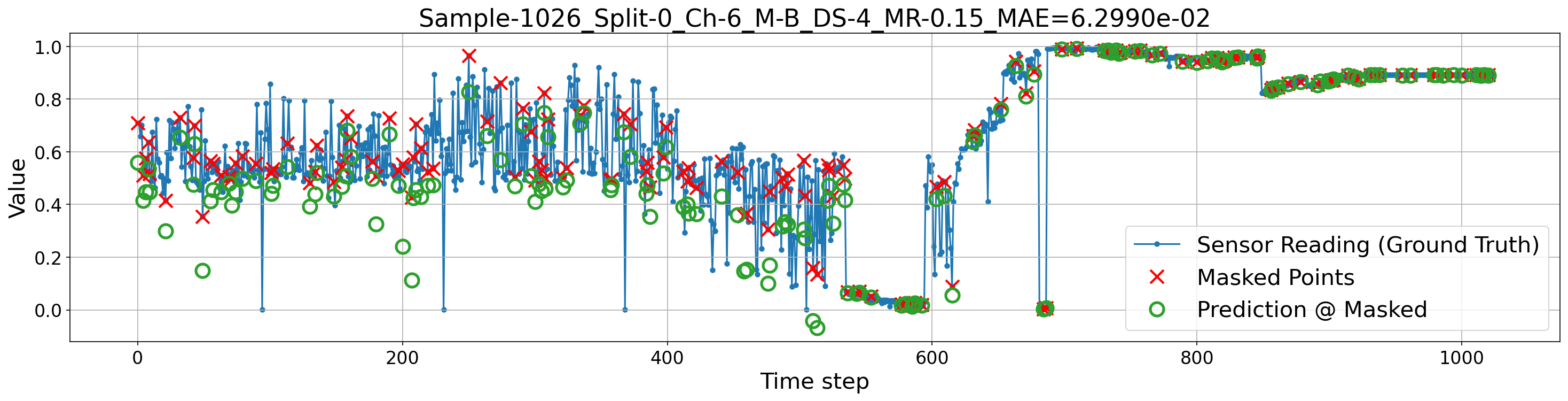}
\end{minipage}
\begin{minipage}{1.0\textwidth}
  \includegraphics[width=0.95\linewidth,trim=0cm 0cm 0cm 0cm,clip]{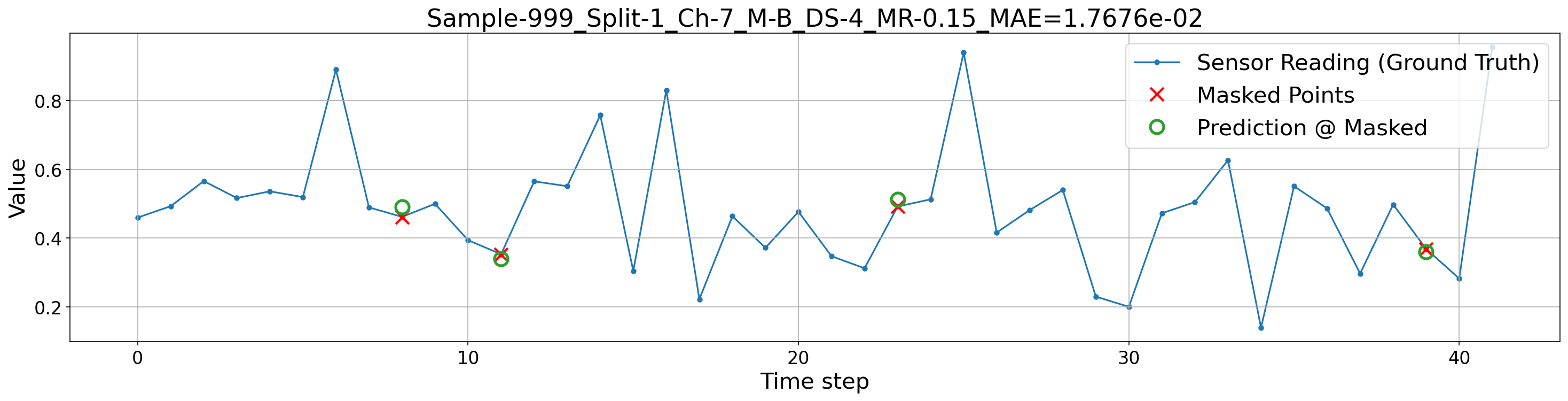}
\end{minipage}
\caption{Representative zero-shot \emph{point-wise imputation} examples from pretrained \emph{SensOFormer-B} on held-out \emph{RFS-LCLS} test windows with a \emph{15\% mask ratio}. Blue curves show the normalized ground-truth sensor values, red markers indicate masked entries, and green circles show the model predictions at the masked locations. Each panel corresponds to a different test window and channel. The subplot title reports the sample index, split, channel, model size (M), dataset (DS), mask ratio (MR), and mean absolute error (MAE) computed on the masked entries.}
\label{fig:viz_mdi_point_ds4}
\end{figure}

\newpage
\subsubsection{Block-wise missing data imputation}
\begin{figure}[h]
\centering
\begin{minipage}{1.0\textwidth}
  \includegraphics[width=0.95\linewidth,trim=0cm 0cm 0cm 0cm,clip]{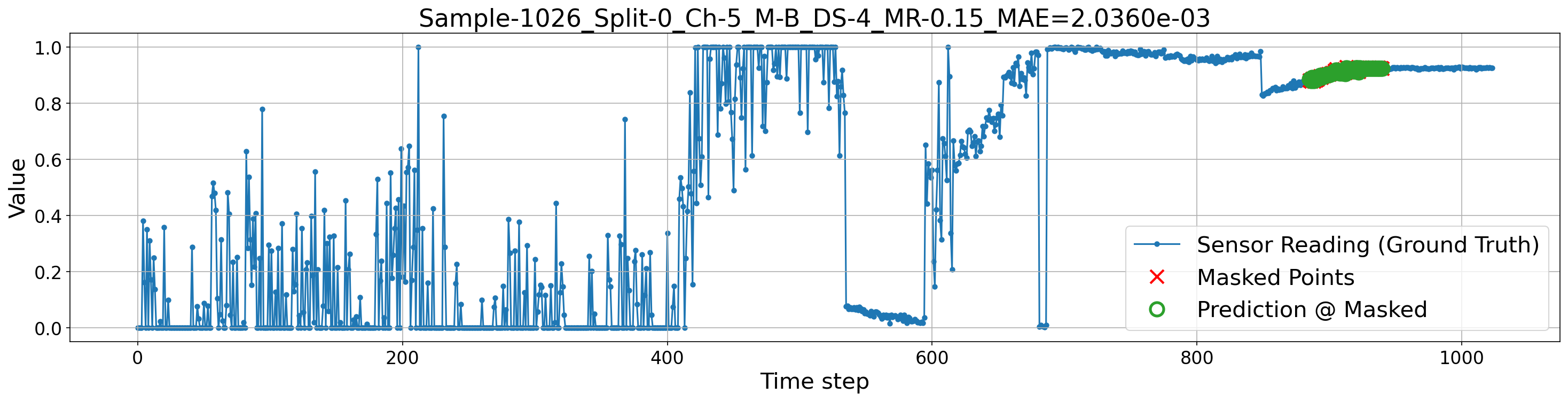}
\end{minipage}
\begin{minipage}{1.0\textwidth}
  \includegraphics[width=0.95\linewidth,trim=0cm 0cm 0cm 0cm,clip]{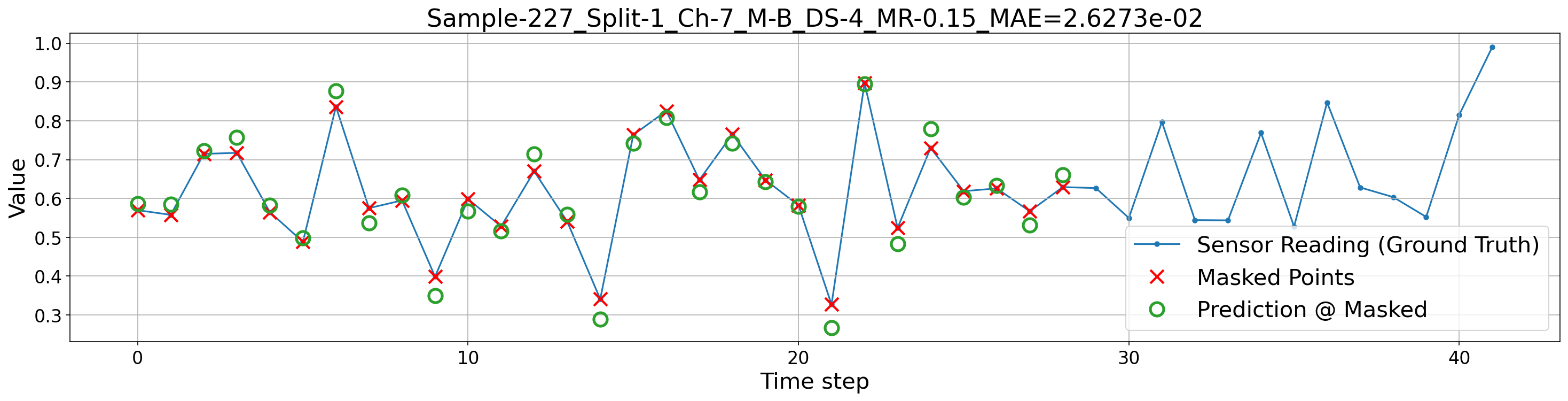}
\end{minipage}
\begin{minipage}{1.0\textwidth}
  \includegraphics[width=0.95\linewidth,trim=0cm 0cm 0cm 0cm,clip]{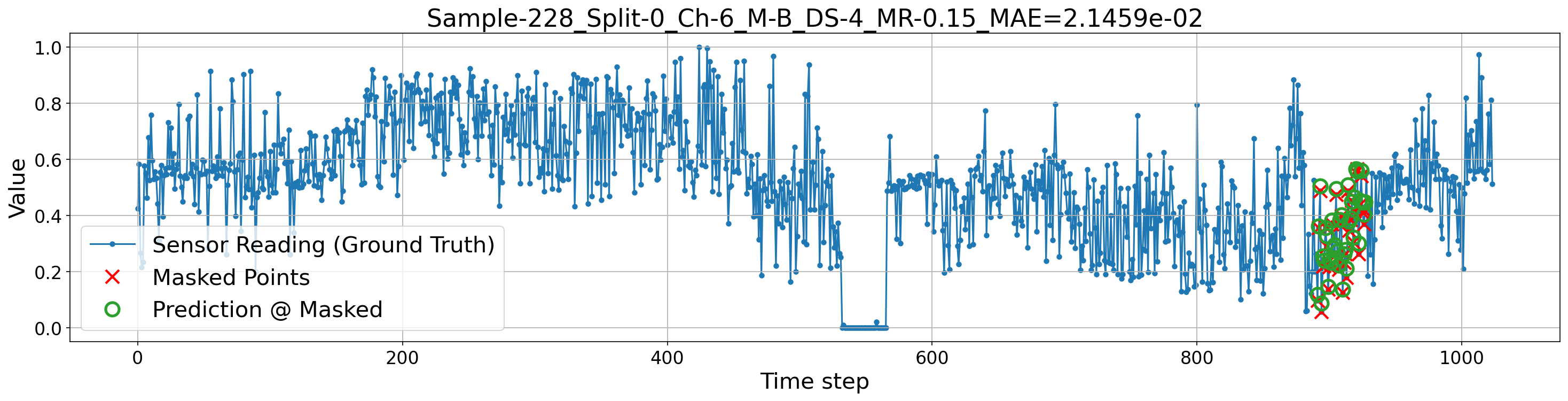}
\end{minipage}
\begin{minipage}{1.0\textwidth}
  \includegraphics[width=0.95\linewidth,trim=0cm 0cm 0cm 0cm,clip]{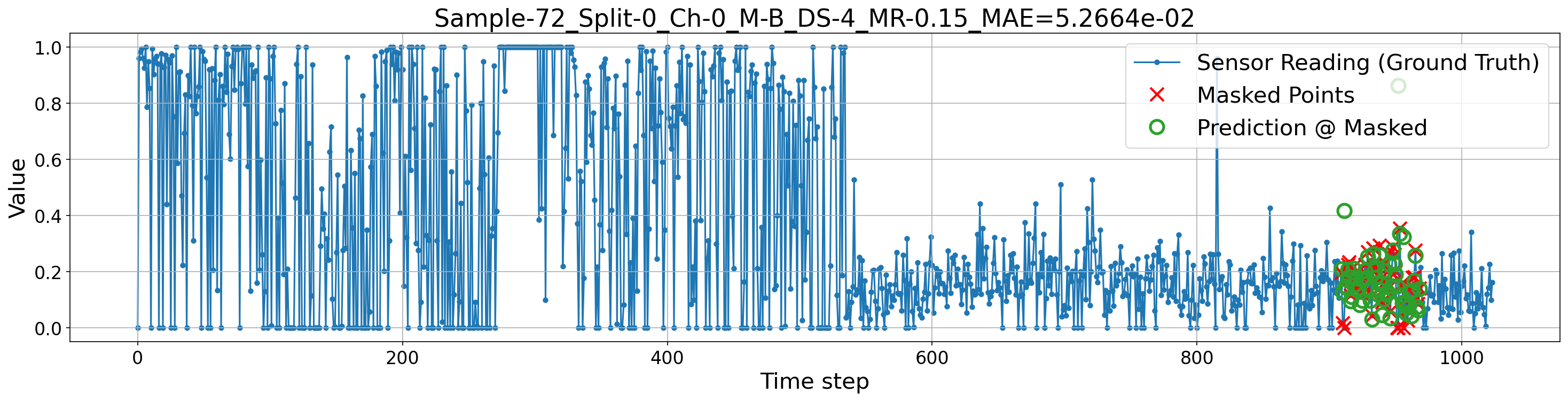}
\end{minipage}
\begin{minipage}{1.0\textwidth}
  \includegraphics[width=0.95\linewidth,trim=0cm 0cm 0cm 0cm,clip]{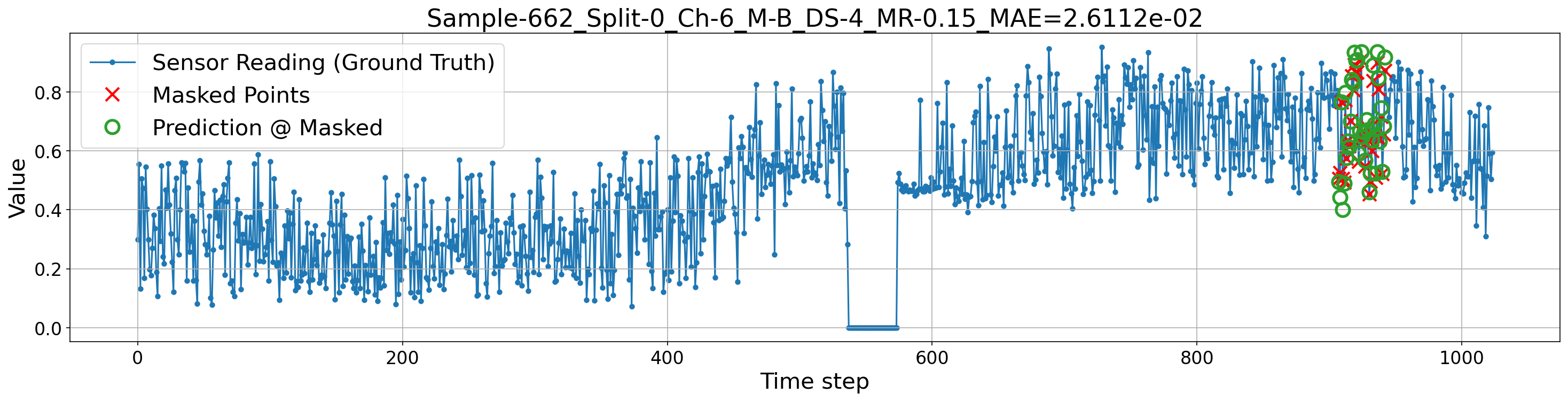}
\end{minipage}
\caption{Representative zero-shot \emph{block-wise imputation} examples from pretrained \emph{SensOFormer-B} on held-out \emph{RFS-LCLS} test windows with a \emph{15\% mask ratio}. Blue curves show the normalized ground-truth sensor values, red markers indicate masked entries, and green circles show the model predictions at the masked locations. Each panel corresponds to a different test window and channel. The subplot title reports the sample index, split, channel, model size (M), dataset (DS), mask ratio (MR), and mean absolute error (MAE) computed on the masked entries.}
\label{fig:viz_mdi_block_ds4}
\end{figure}

\newpage
\subsection{GMPS-Booster-Fermilab}
\subsubsection{Block-wise and point-wise missing data imputation}
\begin{figure}[h]
\centering
\begin{minipage}{1.0\textwidth}
  \includegraphics[width=0.95\linewidth,trim=0cm 0cm 0cm 0cm,clip]{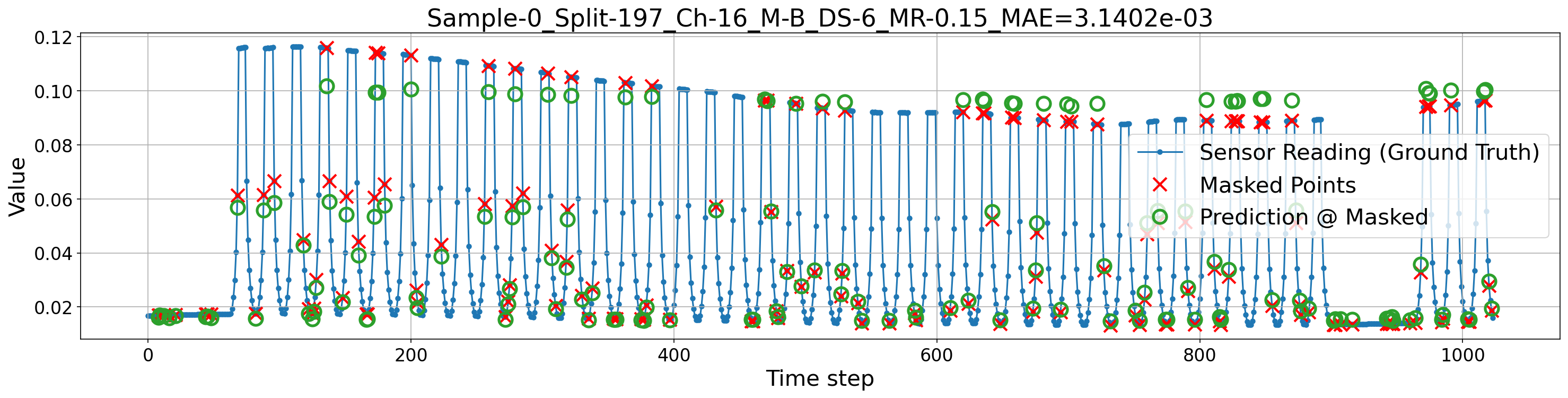}
\end{minipage}
\begin{minipage}{1.0\textwidth}
  \includegraphics[width=0.95\linewidth,trim=0cm 0cm 0cm 0cm,clip]{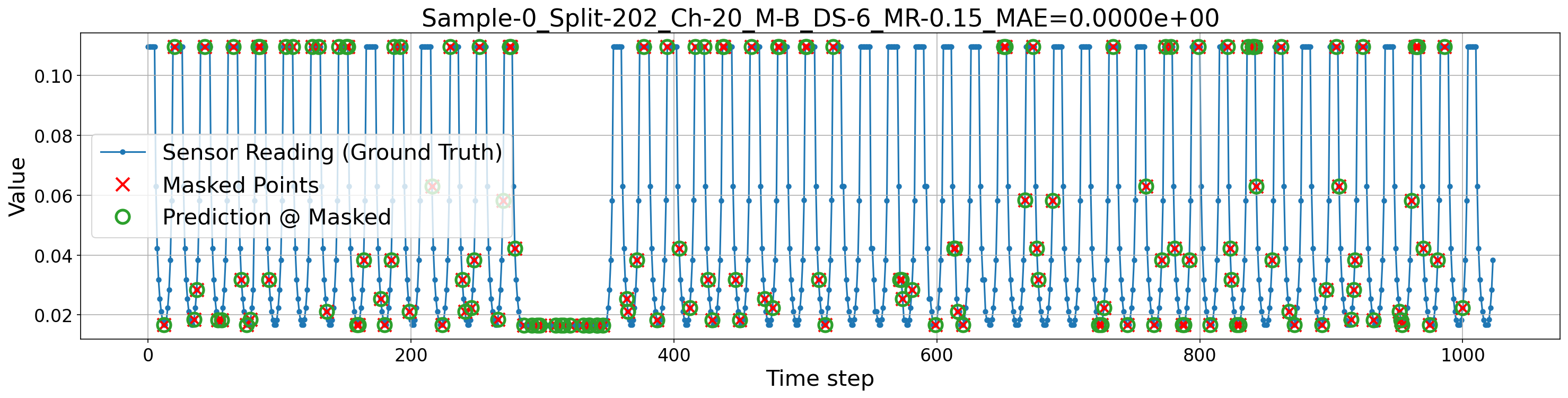}
\end{minipage}
\begin{minipage}{1.0\textwidth}
  \includegraphics[width=0.95\linewidth,trim=0cm 0cm 0cm 0cm,clip]{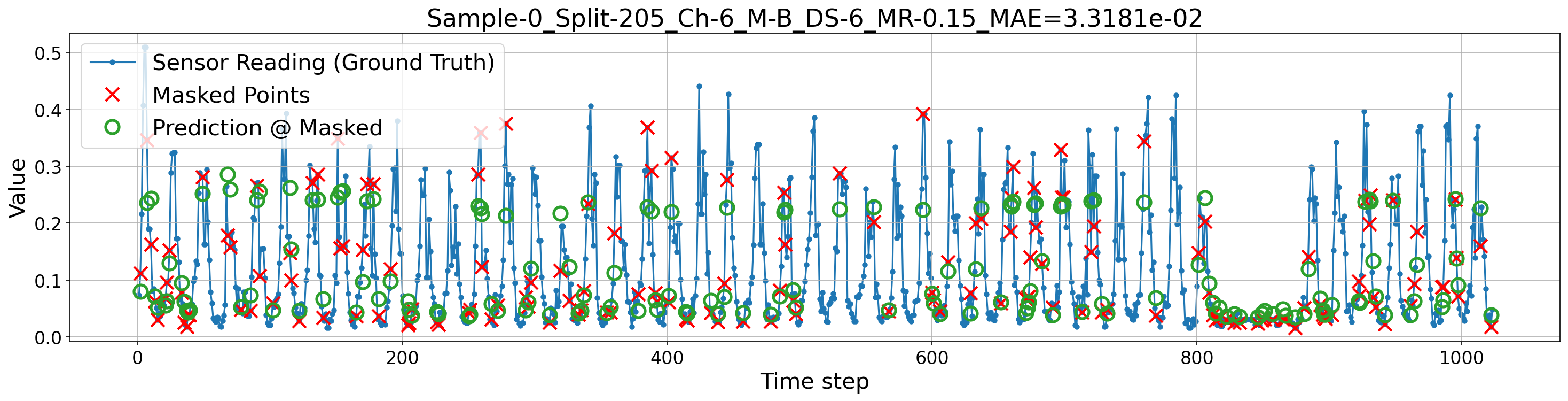}
\end{minipage}
\begin{minipage}{1.0\textwidth}
  \includegraphics[width=0.95\linewidth,trim=0cm 0cm 0cm 0cm,clip]{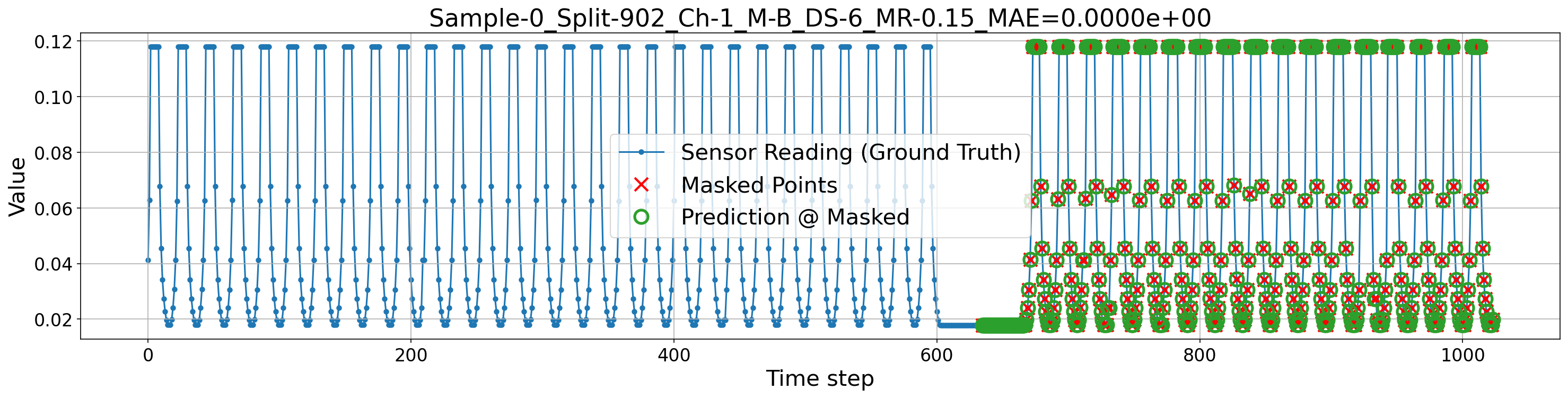}
\end{minipage}
\begin{minipage}{1.0\textwidth}
  \includegraphics[width=0.95\linewidth,trim=0cm 0cm 0cm 0cm,clip]{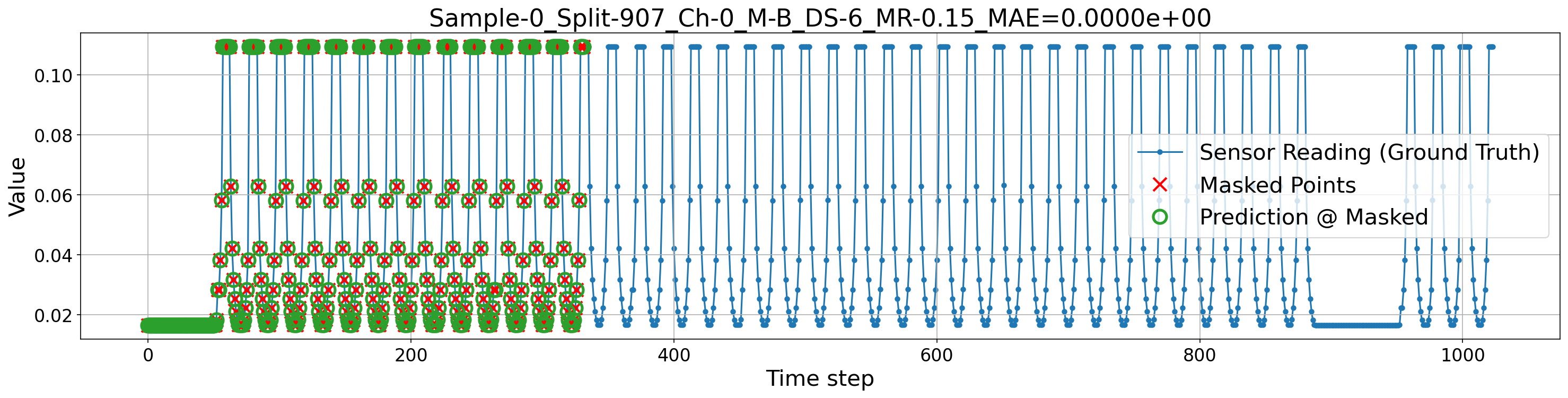}
\end{minipage}
\caption{Representative zero-shot \emph{block-wise and point-wise imputation} examples from pretrained \emph{SensOFormer-B} at the previously unseen \emph{GMPS-Booster-Fermilab} facility with a \emph{15\% mask ratio}. Blue curves show the normalized ground-truth sensor values, red markers indicate masked entries, and green circles show the model predictions at the masked locations. Each panel corresponds to a different test window and channel. The subplot title reports the sample index, split, channel, model size (M), dataset (DS), mask ratio (MR), and mean absolute error (MAE) computed on the masked entries.}
\label{fig:viz_mdi_gmps}
\end{figure}

\end{document}